\documentclass[]{thinking_with_video}

\usepackage[utf8]{inputenc}
\usepackage{threeparttable}
\usepackage{algorithm}
\usepackage{algorithmic}
\usepackage{newfloat}
\usepackage{listings}
\usepackage{makecell}
\usepackage{array}
\usepackage{tabularx}
\usepackage{enumitem}
\usepackage{pifont}
\usepackage{float}
\usepackage{titletoc}
\usepackage{url}

\newcommand\bench{VectorHarness-Bench}
\newcommand{\method}{VectorHarness}

\DeclareCaptionStyle{ruled}{labelfont=normalfont,labelsep=colon,strut=off}
\floatstyle{ruled}
\newfloat{listing}{tb}{lst}{}
\floatname{listing}{Listing}

\title{VectorHarness: Recovering Editable, Relation-Preserving Structure from Scientific Graphics}

\author{
  \mbox{Jiahao Tang\textsuperscript{1}},
  \mbox{Yiren Song\textsuperscript{2}},
  \mbox{Alex Jinpeng Wang\textsuperscript{1,$\S$}}
}

\renewcommand{\affiliationlist}{%
{\affiliationfont
$^{1}$CSU-JPG, Central South University
\qquad
$^{2}$National University of Singapore\\[0.6mm]
}%
}

\abstract{
Converting scientific graphics into editable representations remains a challenging problem for image-to-code generation because of their heterogeneous elements and complex layouts.
Recent multi-agent reconstruction systems have advanced this line of work, but often follow a \textit{copy-paste} paradigm: the reconstructed image closely resembles the original, while complex regions remain effectively uneditable.
We instead formulate a different objective, \textit{raster-to-authoring reconstruction}, which aims to recover an authoring representation that supports native, customized editing rather than mere visual replication.
To this end, we present \textbf{\method{}}, a multi-agent framework for raster-to-authoring reconstruction that recovers heterogeneous components using type-appropriate native representations.
Text, formulas, shapes, connectors, icons, charts, and tables are reconstructed as natively editable objects, while intrinsically image-based regions remain raster content.
To systematically evaluate reconstruction quality, we introduce \textbf{\bench}, which jointly assesses rendering fidelity, raster fallback coverage, executable object edits, and relation-preserving edits. 
Experiments show that \method~ improves executable edit success and relation preservation, reduces avoidable raster fallback, and maintains high visual fidelity across heterogeneous graphics.

}
\checkdata[Repository]{\url{https://github.com/CSU-JPG/Edit_Agent}}

\begin{document}
\maketitle
\renewcommand{\thefootnote}{}
\renewcommand{\thefootnote}{\arabic{footnote}}

\vspace{-1.2em}

\section{Introduction}

A scientific graphic can be reconstructed almost pixel for pixel and still remain effectively uneditable.
Unlike ordinary images, scientific graphics are authoring artifacts composed of heterogeneous, typed elements: text has character structure, formulas have symbolic representations, connectors attach to specific objects, charts and tables encode internal organization, and visual components participate in grouping, containment and hierarchy.
However, figures, diagrams, charts, and posters are often available only as flattened raster images after document extraction, screenshotting, format conversion, or raster-based generation~\cite{hazimeh2025semantic,suzuki2025layerd,chen2026creatiparser}.
Flattening preserves rendered appearance but discards the authoring state required for revision.
Consequently, even local edits---such as correcting a label, updating data, recoloring a shape, rerouting a connector, or rearranging a group---may require substantial manual reconstruction.
\begin{figure}[t]
\centering
\includegraphics[width=1.0\linewidth]{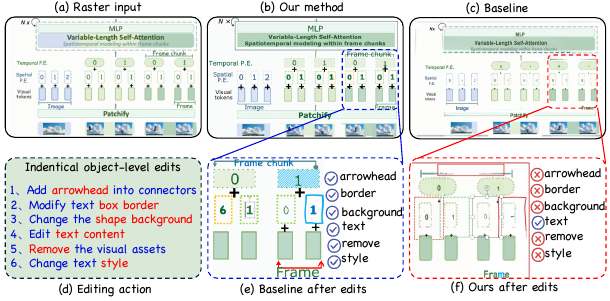}

\caption{
\textit{Copy-Paste} vs. \textit{Native Editability}.
Existing systems~\cite{EditBanana} may preserve complex regions using raster patches or non-native primitives, producing visually similar outputs that do not support the intended editing operations.
In contrast, \method{} reconstructs semantic components as native authoring objects, enabling localized edits to text styles, shape geometry, colors, and connectors while preserving overall visual fidelity.
}
\label{fig:teaser}

\end{figure}

\begin{table}[t]
\centering
\caption{Comparison of supported graphic categories, native recovery, raster policy, and export formats.}
\label{tab:capability_comparison}
\scriptsize
\setlength{\tabcolsep}{3.2pt}
\renewcommand{\arraystretch}{1.08}

\begin{threeparttable}
\begin{tabular}{lccccccc}
\toprule
\textbf{Method} & \textbf{AP} & \textbf{Diag.} & \textbf{DP} & \textbf{Tab.} & \textbf{Chart} & \textbf{Raster} & \textbf{Fmt.} \\
\midrule
AutoFigure-Edit & $\circ$ & $\circ$ & -- & -- & -- & Yes & SVG \\
Edit-Banana     & $\circ$ & $\circ$ & -- & -- & -- & Yes & XML \\
CraftEditor     & $\circ$ & $\circ$ & $\circ$ & -- & -- & Yes & SVG \\
\midrule
\textbf{\method{}} & \checkmark & \checkmark & \checkmark & \checkmark & \checkmark & \textbf{Intrinsic} & \textbf{SVG/XML/PPTX} \\
\bottomrule
\end{tabular}

\begin{tablenotes}[flushleft]
\footnotesize
\item AP: academic poster; DP: design poster; Tab.: table. 
$\checkmark$: native editable recovery; 
$\circ$: editable output with raster fallback for some recoverable elements; 
Intrinsic: raster retained only for intrinsically raster content.
\end{tablenotes}
\end{threeparttable}

\end{table}
This limitation reflects a mismatch between reconstruction objectives and authoring requirements.
Classical vectorization converts raster observations into contours, B\'ezier paths, or layered geometric primitives~\cite{selinger2003potrace,li2020diffvg,reddy2021im2vec,ma2022live}.
Recent vision-language systems generate SVG code or recover editable visual assets~\cite{rodriguez2025starvector,hazimeh2025semantic,he2026vfig,huang2026scifig}, while graphic-design parsers decompose images into visual layers~\cite{suzuki2025layerd,chen2026creatiparser}.
Domain-specific methods recover content types, such as formulas as markup or charts as structured data~\cite{deng2017im2markup,liu2023deplot,luo2021chartocr}.
These approaches provide parts of the solution, but typically adopt a single target abstraction or recover one class of structure in isolation.
Whole-graphic reconstruction instead requires heterogeneous representations to coexist within one editing-consistent state: formulas should remain symbolic, connectors should preserve endpoint attachment, charts and tables should expose their internal organization, parameterized shapes should remain adjustable, and intrinsically image-based regions should remain raster.

We therefore formulate \textbf{raster-to-authoring reconstruction}: recovering an editing-consistent authoring representation from a flattened raster graphic rather than merely producing a visually similar rendering.
The reconstruction target is not a particular output format, and editability is not defined by vector paths or syntactically valid code.
Instead, each recoverable component should be instantiated using a representation appropriate to its semantic role and intended editing operations.
A successful reconstruction must preserve component identity, type-specific editable content, intra-object structure, cross-object relations, and rendering fidelity, while retaining raster content only for intrinsically image-based regions.
Table~\ref{tab:capability_comparison} summarizes how these requirements distinguish our formulation from existing reconstruction systems.
Output formats such as SVG, diagrams.net XML, and PPTX are therefore treated as compilation targets rather than the reconstruction state itself.

To address this objective, we present \textbf{\method{}}, a multi-agent framework built around a shared \textbf{Semantic Authoring Representation (SAR)}. A graphic-level router selects a reconstruction branch for scientific diagrams, academic posters, design posters, charts, or tables, where specialized agents recover heterogeneous components in type-appropriate representations. The recovered objects are assembled into SAR together with operational relations, including grouping, containment, association, alignment, hierarchy, and connector attachment. Instead of relying solely on a final visual-quality critic, stage-local verification evaluates object validity, relation consistency, and rendering fidelity, routing localized failures to the responsible stages for correction without regenerating validated content. Finally, deterministic format-specific compilers translate the same SAR into layered SVG, diagrams.net-compatible XML, and editable PowerPoint objects, supporting consistent authoring across multiple output environments.

Raster-to-authoring reconstruction also requires an evaluation protocol that measures more than whether an output looks correct.
Existing evaluations commonly emphasize rendered similarity, code validity, structural validity, or holistic visual preference~\cite{rodriguez2025starvector,hazimeh2025semantic,he2026vfig,zhao2026crafter,mao2026residual,mao2026textground4m,song2024diffsim}. Task-specific multimodal benchmarks likewise demonstrate the importance of measuring whether visual systems satisfy operational criteria beyond perceptual plausibility~\cite{wu2025mcabench}
These metrics cannot reliably distinguish a native authoring object from a raster patch or fragmented primitive that visually imitates it.
We therefore introduce \textbf{\bench}, spanning scientific diagrams, charts, tables, academic posters, and design posters.
The benchmark jointly evaluates rendering fidelity, raster fallback coverage, executable object edits, and relation-preserving edits.
Experiments show that \method{} improves executable edit success and relation preservation, reduces avoidable raster fallback, and maintains high visual fidelity across heterogeneous graphic categories and output formats.

Our contributions are summarized as follows:

\begin{itemize}

\item We formulate \textbf{raster-to-authoring reconstruction}, distinguishing editing-consistent recovery from conventional visual replication through typed native representations and cross-object relations.

\item We propose \textbf{\method{}}, a multi-agent framework built around a shared Semantic Authoring Representation, enabling representation-aware component recovery, relation-aware assembly, stage-local verification, and deterministic compilation into SVG, XML, and PPTX.

\item We introduce \textbf{\bench}, an evaluation protocol that jointly measures rendering fidelity, raster fallback coverage, executable object edits, and relation-preserving edits across heterogeneous scientific graphics.

\end{itemize}

\begin{figure*}[t]
\centering
\includegraphics[width=\textwidth]{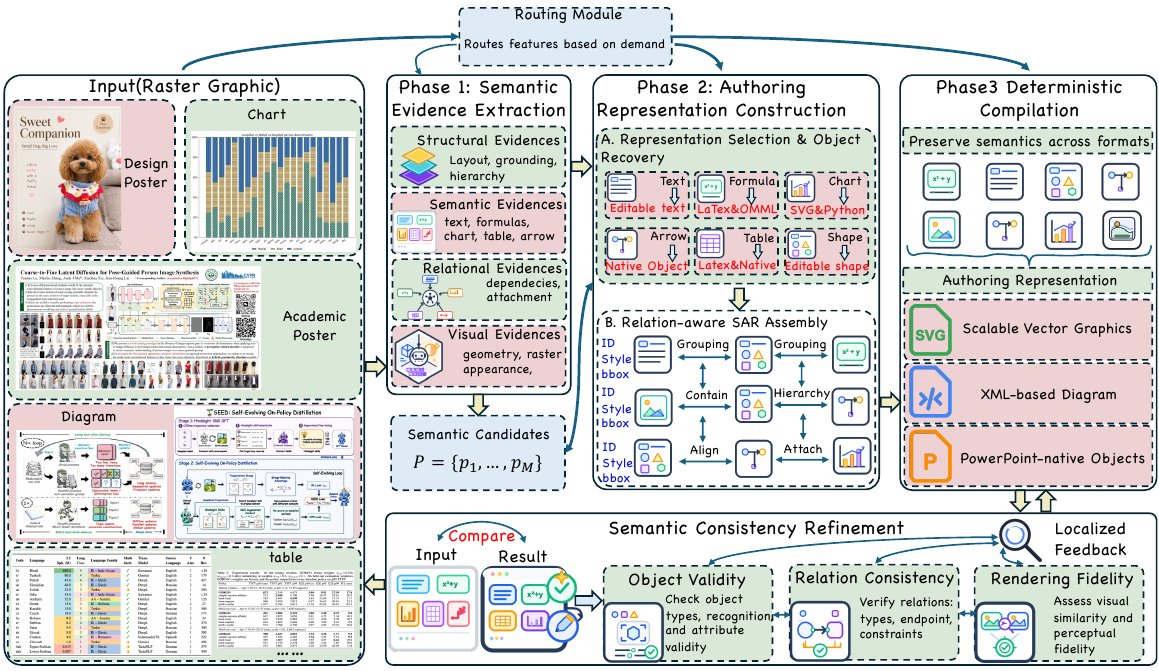}

\caption{
Overview of \method{}. Given a flattened raster graphic, \method{} first analyzes its global layout and routes the input to an appropriate reconstruction branch. Specialized parsing and recovery agents then reconstruct text, formulas, shapes, connectors, icons, charts, tables, and raster regions using type-appropriate representations. The recovered objects and their grouping, containment, association, alignment, hierarchy, and attachment relations are assembled into a shared Semantic Authoring Representation (SAR). Stage-local verification checks object validity, relation consistency, and rendering fidelity, and routes localized errors back for targeted repair. Finally, deterministic format-specific compilers translate the verified SAR into layered SVG, diagrams.net-compatible XML, and editable PPTX outputs.
}
\label{fig:overview}

\end{figure*}

\section{Related Work}

\subsection{Vector graphics generation.}

Classical raster-to-vector methods trace image contours and approximate them using paths or geometric primitives, as exemplified by Potrace~\cite{selinger2003potrace}. Learning-based approaches model vector programs or optimize primitives through differentiable rendering, including DeepSVG~\cite{carlier2020deepsvg}, DiffVG~\cite{li2020diffvg}, Im2Vec~\cite{reddy2021im2vec}, and LIVE~\cite{ma2022live}, and SuperSVG~\cite{hu2024supersvg}.
More recent methods leverage diffusion models or vision-language models for text- and image-conditioned SVG generation. 
VectorFusion~\cite{jain2023vectorfusion} and SVGDreamer~\cite{xing2024svgdreamer} optimize vector graphics under pretrained image-generation priors, while StarVector~\cite{rodriguez2025starvector} and Chat2SVG~\cite{wu2025chat2svg} generate SVG code using multimodal or language-based models. Related work explores layered SVG synthesis, differentiable vector-graphics editing, vector WordArt generation, and text-driven appearance synthesis~\cite{song2025layertracer,song2023clipvg,song2022clipfont,song2022cliptexture}. 
These methods advance scalable vector generation and geometric reconstruction, but primarily target paths, SVG primitives, or complete SVG programs. 
They are not designed to recover the heterogeneous authoring semantics of scientific graphics, such as symbolic formulas, data-backed charts, table topology, connector attachment, and cross-object grouping.

\subsection{Structured Visual Reconstruction}

Prior work recovers structured representations from specific raster graphics. Draw with Thought~\cite{cui2025draw} reconstructs scientific diagrams as editable mxGraph XML, while VFIG~\cite{he2026vfig} learns figure-to-SVG conversion with vision-language models. For charts, ChartCoder~\cite{zhao2025chartcoder} and CharTide~\cite{zheng2026chartide} improve chart-to-code generation, while Chart2Code~\cite{tang2026chartscodehierarchicalbenchmark} evaluates chart reproduction, editing, and table-to-chart generation. Other systems recover formulas, chart data, or editable design layers~\cite{deng2017im2markup,liu2023deplot,luo2021chartocr,suzuki2025layerd,chen2026creatiparser}. Complementary work studies layout generation, layer-wise alpha decomposition, relation transfer, and controllable text or typography rendering~\cite{wan2024grid,wang2025diffdecompose,gong2025relationadapter,lu2025easytext,shi2024fonts,shi2025wordcon}. These methods remain specialized by graphic category, object type, or output format. In contrast, we recover a unified, typed, and relation-aware authoring state spanning text, formulas, charts, tables, connectors, primitives, icons, groups, and raster regions.

\subsection{Scientific Figure Vectorization}

Recent agentic systems generate and edit scientific illustrations. PaperBanana~\cite{zhu2026paperbanana} generates illustrations from scientific content, while AutoFigure-Edit~\cite{lin2026autofigure} and LiveFigure~\cite{shao2026livefigure} produce editable SVG or presentation-native figures. Broader work on customized instruction-based editing, style-agnostic consistency, and procedural sequence generation likewise informs the need to preserve intended visual structure during synthesis and revision~\cite{huang2025arteditor,song2025omniconsistency,song2025makeanything}. Unlike these generation-oriented systems, our setting recovers the authoring state of an existing raster graphic. More closely related, Edit-Banana~\cite{EditBanana} reconstructs editable diagram representations, while CraftEditor, introduced as part of Crafter~\cite{zhao2026crafter} converts raster figures into editable SVGs through decomposition and iterative refinement. However, both target particular output representations. \method{} instead recovers a format-agnostic Semantic Authoring Representation containing typed objects and operational relations, then deterministically compiles it into layered SVG,  XML, and PPTX.


\section{Method}
\label{sec:method}

\subsection{Problem Formulation}
\label{sec:problem}

Given a flattened raster graphic $I$, \method{} reconstructs an editable artifact by first recovering a format-agnostic authoring state and then compiling it into one or more target formats. 
We formulate this problem as \emph{raster-to-authoring reconstruction}:
\begin{equation}
\mathcal{A}=F(I), \qquad V_m=C_m(\mathcal{A}),
\label{eq:formulation}
\end{equation}
where $\mathcal{A}$ is the recovered authoring representation, $F$ is the multi-agent reconstruction process, and $C_m$ is a deterministic compiler for target format $m$.

This formulation treats layered SVG, diagrams.net XML, and PPTX as compilation targets rather than reconstruction states.
A visually faithful output may remain difficult to edit when text is converted into glyph outlines, formulas into unrelated paths, connectors into detached line segments, or structured graphics into raster patches. 
We therefore seek to preserve semantic identity, type-specific editable content, operational relations, and rendering fidelity.

\method{} consists of four stages, as illustrated in Figure~\ref{fig:overview}. 
A graphic-level router selects a parsing branch; representation-aware agents recover heterogeneous objects; relation-aware assembly consolidates them into the Semantic Authoring Representation; and deterministic compilers generate editable outputs. 
Stage-local verification identifies errors and routes them to the responsible agent. 
Implementation details, including agent configurations, prompts, validation rules, and format-specific mappings, are provided in the appendix.

\subsection{Semantic Authoring Representation}
\label{sec:sar}
Different components require different authoring representations. 
Text should preserve character content and typography; formulas should retain symbolic structure; connectors should maintain endpoint dependencies; and intrinsically image-based regions should preserve their raster content. 
A uniform representation based solely on paths or image layers cannot retain these type-specific editing operations.

We introduce the \textbf{Semantic Authoring Representation (SAR)}, a typed and relation-aware intermediate stage:
\begin{equation}
\mathcal{A}=(O,E),
\label{eq:sar}
\end{equation}
where $O=\{o_i\}_{i=1}^{N}$ contains reconstructed object records and $E$ contains explicit relation records. 
Each object is represented as
\begin{equation}
o_i=(\tau_i,r_i,c_i,b_i,g_i,a_i,z_i),
\label{eq:object}
\end{equation}
where $\tau_i$ is the semantic type, $r_i$ is the selected authoring representation, $c_i$ is the type-specific editable payload, $b_i$ is the spatial support, $g_i$ and $a_i$ describe geometry and appearance, and $z_i$ specifies rendering order.

The distinction between semantic type $\tau_i$ and authoring representation $r_i$ is central to our formulation.
The former specifies \emph{what} an object represents, whereas the latter determines \emph{how} it can be edited.
Identifying a region as a formula, for example, does not determine whether it is recovered as a symbolic expression, vectorized glyphs, or a raster crop. 
Similarly, detecting an arrow does not guarantee its reconstruction as a functional connector with attached endpoints.

The payload $c_i$ stores the editable state appropriate to the selected representation. 
Depending on the object type, it may contain text and typographic attributes, symbolic expressions, chart data and visual encodings, table cells and topology, connector endpoints, primitive parameters, vector geometry, or raster masks. 
This shared outer schema allows heterogeneous objects to coexist without reducing their internal semantics to a uniform low-level representation.

The relation set $E$ records dependencies that affect authoring behavior, including grouping, containment, hierarchy, association, alignment, and attachment. 
These relations allow composite elements to be manipulated as semantic units and preserve operational dependencies, such as connector attachment when related objects are moved. 
SAR therefore captures both intra-object editable structure and inter-object organization in a shared state used throughout reconstruction, verification, and compilation.

\subsection{Representation-Aware Reconstruction}
\label{sec:reconstruction}

The central challenge is not merely to detect visible components, but to determine which authoring representation best preserves the editing behavior implied by each component.
To address this challenge, \method{} factorizes the reconstruction process $F$ into graphic routing and evidence parsing, representation-aware object recovery, and relation-aware SAR assembly. 
Specialized agents may use different recognition models and recovery operators, while their outputs are normalized into the shared SAR schema.
\paragraph{Graphic routing and evidence parsing.}

A graphic-level router selects a reconstruction branch suited to the dominant structure of the input. 
The current system provides specialized branches for scientific diagrams, academic and design posters, charts, tables, and image-centric compositions. 
This routing strategy allows each graphic category to employ appropriate parsing and recovery operators while retaining a unified downstream representation.

The selected parsing branch decomposes the input into regions and extracts four forms of evidence.
Structural evidence describes panels, layout, hierarchy, grouping cues, and reading order.
Semantic evidence characterizes textual, symbolic, graphical, and data-bearing content.
Relational evidence proposes alignment, containment, association, and attachment dependencies.
Visual evidence records geometry, appearance, masks, and cues indicating whether raster preservation is appropriate.
Unlike parsers that produce only bounding boxes and category labels, this stage collects the evidence needed to determine both \emph{what} each component represents and \emph{how} it should be instantiated for editing.

\paragraph{Representation selection and object recovery.}

For each candidate $p_i$, \method{} jointly selects its semantic type and an admissible authoring representation:
\begin{equation}
(\tau_i,r_i)=
\Phi(p_i,I,\mathcal{C}_i),
\qquad r_i\in\mathcal{R}(\tau_i),
\label{eq}
\end{equation}
where $\mathcal{C}_i$ contains contextual evidence from neighboring candidates, structural predictions, relation proposals, and the current SAR state, and $\mathcal{R}(\tau_i)$ denotes the set of representations admissible for semantic type $\tau_i$.

The selection operator $\Phi$ favors the most structured representation reliably supported by the available evidence, rather than the representation that most directly reproduces visible pixels. 
For example, vectorized glyphs may approximate the appearance of a formula, but a symbolic expression preserves its mathematical editing operations. 
Similarly, an isolated line path may resemble an arrow while lacking the endpoint dependencies required of a functional connector. 
This separation between semantic recognition and authoring instantiation distinguishes raster-to-authoring reconstruction from appearance-oriented derendering.

The selected pair $(\tau_i,r_i)$ activates a type-specific recovery operator that produces the editable payload $c_i$ with spatial support, geometry, appearance, and rendering order. 
Text retains character content and typography, while formulas preserve symbolic expressions when supported by visual evidence. 
Charts expose recoverable data and encodings, and tables retain cell content and row--column structure. 
Connectors preserve geometry, endpoints, arrowheads, and attachments. Parameterized shapes, icons, and composite elements are recovered as editable objects or structured groups, while intrinsically image-based regions remain raster.

Components with recoverable internal structure are instantiated as authoring objects. Photographs, microscopy panels, continuous-tone illustrations, and intrinsically image-based regions remain positioned and masked raster objects. This evidence-aware policy avoids both inappropriate over-vectorization and unnecessary raster fallback.

\paragraph{Relation-aware assembly.}

Objects recovered independently do not by themselves form a coherent authoring state.
\method{} therefore consolidates them into SAR by resolving rendering order and instantiating cross-object relations in $E$ from structural evidence, agent proposals, and component-specific constraints.

Relation inference combines spatial compatibility, semantic compatibility, and type-specific structural rules. 
The resulting relations are operational rather than merely descriptive. 
Attachment relations connect functional connector endpoints to their source and target objects. 
Grouping relations allow composite components to be moved or resized as semantic units. Containment and hierarchy encode nested organization, while association relations link labels, legends, and annotations to the content they describe.

Assembly additionally checks that every relation refers to existing and semantically compatible objects and resolves conflicting proposals before they enter SAR. The resulting representation combines independently recovered components into an editing-consistent state that can be locally verified and compiled across target formats.

\subsection{Stage-Local Verification and Repair}
\label{sec:verification}

Raster-to-authoring reconstruction is ambiguous because visually similar renderings may correspond to authoring representations with different editing capabilities. \method{} therefore performs verification within the parsing, object-recovery, and relation-assembly stages rather than relying solely on a final rendering-level critic.

Verification considers three complementary aspects. \emph{Object validity} checks whether an object's semantic type, selected representation, editable payload, geometry, and appearance are mutually consistent. In particular, it verifies that the representation retains the editable degrees of freedom implied by the object's semantic role. \emph{Relation consistency} validates grouping, containment, association, hierarchy, alignment, and attachment dependencies. \emph{Rendering fidelity} compares provisional renderings with the input and identifies discrepancies in content, layout, typography, geometry, appearance, and local composition.

At refinement step $t$, a stage-local verifier produces a structured error record $e^{(t)}$ that identifies the responsible stage $s^{(t)}$ and the affected SAR entries $S^{(t)}$. The error is routed to the corresponding parsing, recovery, or assembly operator, and only the affected subset is revised:
\begin{equation}
\mathcal A^{t+1}_{S^{(t)}} =
U_{s^{(t)}}(I,\mathcal A^t,e^t),\quad
\mathcal A^{t+1}_{\bar S^{(t)}} =
\mathcal A^t_{\bar S^{(t)}}.
\label{eq:localized_repair}
\end{equation}
where $U_{s^{(t)}}$ denotes the operator responsible for the failure, and $\bar{S}^{(t)}$ denotes the unaffected entries. Validated objects and relations are retained rather than regenerated.

Refinement terminates when no actionable error remains or the predefined iteration budget is reached. This localized strategy limits unintended changes to validated content while improving representation correctness and rendering fidelity.

\subsection{Deterministic Multi-Format Compilation}
\label{sec:compilation}

After verification, a deterministic compiler $C_m$ maps SAR objects, payloads, relations, and rendering order to the primitives supported by target format $m$. Separating reconstruction from serialization allows the same authoring state to be reused across multiple environments and avoids requiring a generative model to directly produce complete format-specific documents such as SVG, XML, or PPTX.

Whenever the target format provides a native counterpart, the compiler emits the corresponding editable primitive. Text is serialized as editable text, parameterized geometry as native shapes with adjustable parameters, groups as structured containers, and connectors as attached edges in formats that support endpoint references. When no direct counterpart exists, the compiler constructs a structured composition of editable primitives and preserves the corresponding organization and relations to the extent supported by the target environment. Intrinsically image-based regions remain positioned and masked raster objects.

\method{} currently provides compilation backends for layered SVG, diagrams.net-compatible XML, and editable PPTX. Although these formats differ in their available primitives and relation semantics, all outputs are derived from the same SAR and therefore retain a consistent recovered organization. Extending the system to another authoring environment requires a new compiler backend and format-specific mappings, without repeating visual parsing or semantic reconstruction. SAR thus serves as the single reconstruction state, while individual output formats provide different authoring views of the recovered graphic.

\section{Experiments}
\label{sec:experiments}

We evaluate whether \method{} can jointly preserve visual fidelity and native editability when reconstructing flattened graphics. 
Since a raster input does not provide a unique ground-truth object hierarchy or authoring structure, we do not rely on exhaustive object- or relation-level annotations. 
Instead, our evaluation focuses on directly observable properties of the reconstructed artifacts: rendered fidelity, textual consistency, raster fallback, executable editing operations, and practical human editability. 
We further study the effects of representation selection and
stage-local verification.

\subsection{Experimental Setup}
\label{sec}

\paragraph{Benchmark.}

\begin{figure}[t]
\centering
\includegraphics[width=1\linewidth]{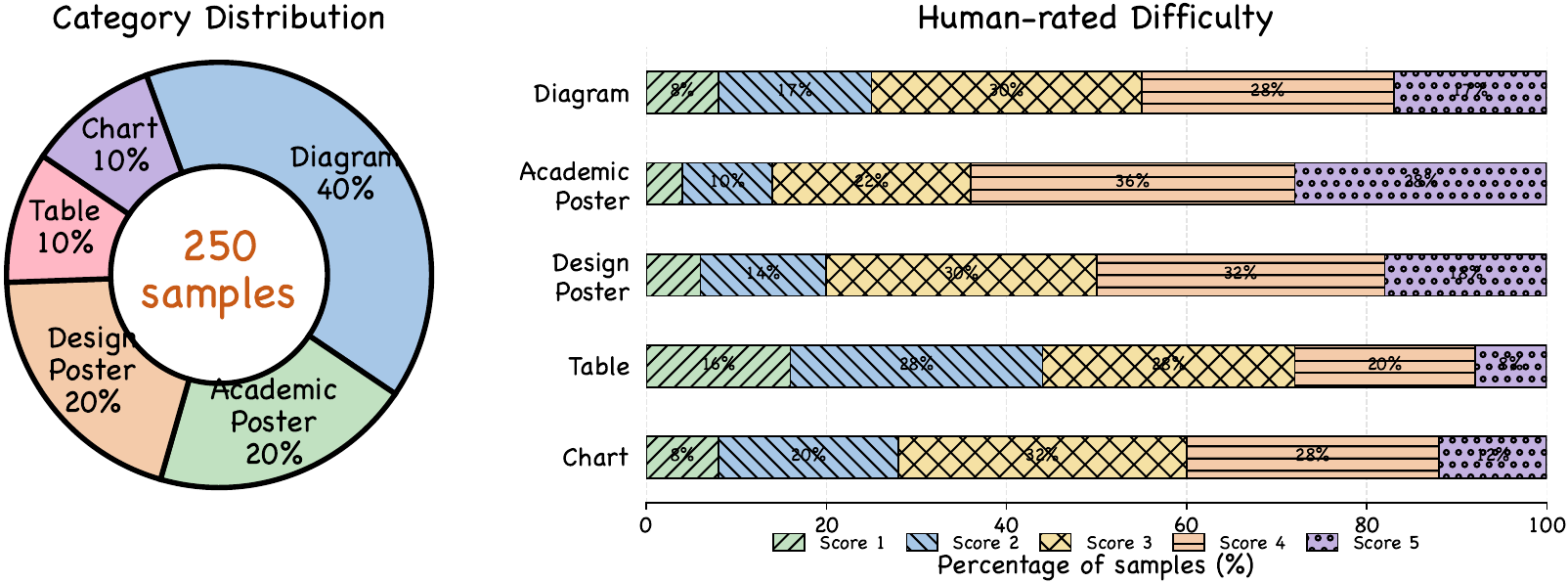}
\caption{
Benchmark composition and human-rated difficulty. Left: Category distribution of 250 graphics across five categories. Right: Category-wise distribution of difficulty scores from 1 (easiest) to 5 (hardest).
}
\label{fig:benchmark}

\end{figure}

We construct \textbf{\bench{}}, containing 250 flattened graphics across five categories: scientific diagrams, charts, tables, academic posters, and design posters. Its composition and difficulty distribution are shown in Figure~\ref{fig:benchmark}. The benchmark covers text, formulas, shapes, connectors, icons, charts, tables, and raster panels. Dataset sources, category statistics, and collection procedures are provided in the appendix.

\paragraph{Baselines.}

We compare \method{} with three task-aligned raster-to-editable systems: \textbf{Edit-Banana}, \textbf{AutoFigure-Edit}, and \textbf{CraftEditor}. We also include a strong \textbf{Direct MLLM} baseline (GPT-5.5), which generates editable SVG without specialized routing, representation selection, or stage-local verification. Additional contour-tracing and multimodal baselines are reported in the appendix.

\paragraph{Evaluation protocol.}
All methods receive the same raster and canvas size without source files or manual correction. We use official output formats and recommended settings, standardize rendering, and evaluate authoring operations on native SVG, diagrams.net XML, or PPTX files. Unsupported categories count as failures in the all-category result; common-domain results are reported in the appendix. Dataset construction, model versions, prompts, parsers, rendering settings, invalid-output handling, confidence intervals, runtime, and cost are also provided in the appendix.

\paragraph{Metrics.}
For automatic evaluation, LPIPS measures perceptual similarity between the input and reconstructed rendering; VLM-Fid. assesses visible-content completeness, layout, and appearance under a fixed VLM protocol; and Text-CER measures textual consistency between OCR outputs from both renderings. \textbf{Raster Fallback Coverage (RFC)} measures the foreground fraction contributed by embedded raster objects on the vector-dominant subset. \textbf{Automatic Edit Success Rate (Auto Edit-SR)} measures whether object edits can be written back while preserving artifact validity and localizing the intended change. \textbf{Relation-Preserving Edit Success Rate (Rel-SR)} measures whether operational relations, including attachment, grouping, and containment, remain valid after editing. For human evaluation, \textbf{Fid.} is the mean visual-fidelity rating, \textbf{Human Edit-SR} is the proportion of tasks successfully completed, and \textbf{Ease} measures perceived editing ease. Full definitions, protocols, and format-specific checkers are provided in the appendix.

\begin{figure*}[t]
\centering
\includegraphics[width=\textwidth]{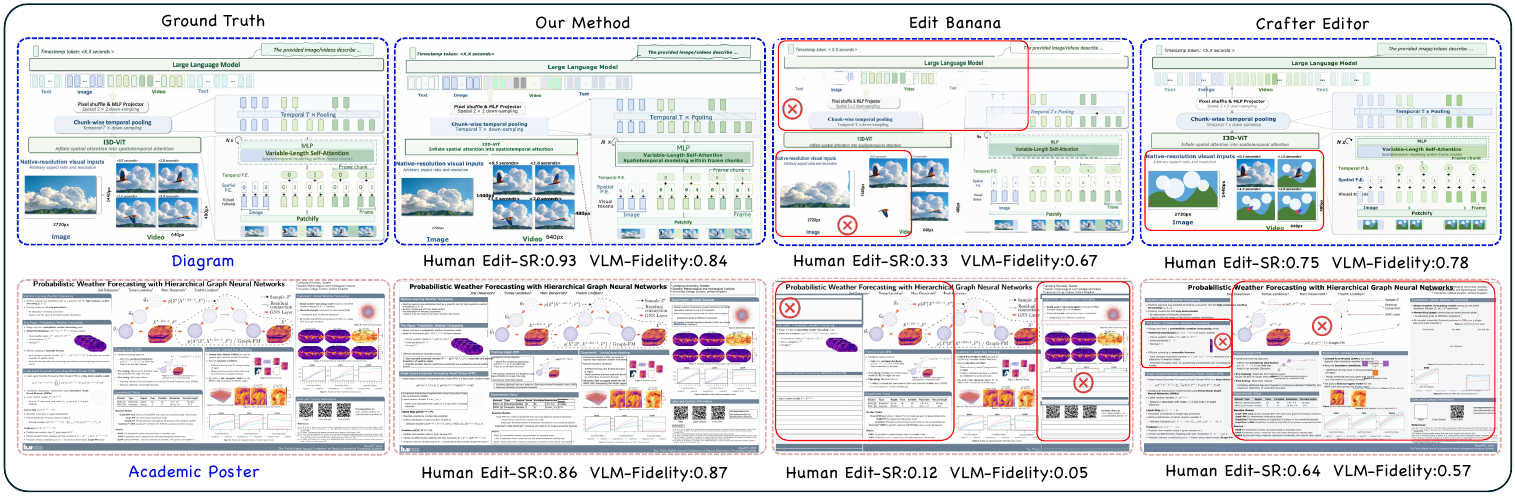}
\caption{
Qualitative comparison on complex diagrams and academic posters. Existing methods may leave text, formulas, arrows, or shapes as raster patches or fragmented primitives, limiting native editing operations. In contrast, \method{} reconstructs recoverable visual components as native editable objects, enabling localized edits while maintaining high visual fidelity.
}
\label{fig:qualitative_results}

\end{figure*}

\begin{table*}[t]
\centering
\footnotesize
\setlength{\tabcolsep}{3.2pt}
\renewcommand{\arraystretch}{1.10}

\begin{tabular}{l|ccc|ccc|ccc}
\toprule
\textbf{Method}
&
\multicolumn{3}{c|}{\textbf{Fidelity}}
&
\multicolumn{3}{c|}{\textbf{Automatic Editability}}
&
\multicolumn{3}{c}{\textbf{Human Evaluation}}
\\

&
\textbf{LPIPS}$\downarrow$
&
\textbf{VLM-Fid.}$\uparrow$
&
\textbf{Text-CER}$\downarrow$
&
\textbf{RFC}$\downarrow$
&
\textbf{Edit-SR}$\uparrow$
&
\textbf{Rel-SR}$\uparrow$
&
\textbf{Fid.}$\uparrow$
&
\textbf{Edit-SR}$\uparrow$
&
\textbf{Ease}$\uparrow$
\\

\midrule

Edit-Banana
& 0.61 & 0.63 & 0.30 & 0.45 & 0.60 & 0.53 & 0.55 & 0.25 & 0.35 \\

AutoFigure-Edit
& 0.56 & 0.67 & 0.21 & 0.37 & 0.69 & 0.42 & 0.51 & 0.36 & 0.42 \\

CraftEditor
& 0.48 & 0.71 & \textbf{0.06} & 0.21 & 0.72 & 0.64 & 0.63 & 0.67 & 0.58 \\

Direct MLLM
& 0.85 & 0.26 & 0.62 & 0.50 & 0.41 & 0.24 & 0.15 & 0.18 & 0.14 \\

\midrule

\textbf{\method{}}
& \textbf{0.41}
& \textbf{0.76}
& 0.12
& \textbf{0.13}
& \textbf{0.79}
& \textbf{0.88}
& \textbf{0.72}
& \textbf{0.76}
& \textbf{0.66}
\\

\midrule

\hspace{2mm}w/o Representation Selection
& 0.51 & 0.36 & 0.14 & 0.42 & 0.57 & 0.73 & 0.53 & 0.51 & 0.37 \\

\hspace{2mm}w/o  Stage-Local Verification
& 0.44 & 0.64 & 0.12 & 0.17 & 0.63 & 0.64 & 0.65 & 0.73 & 0.67 \\

\bottomrule
\end{tabular}

\caption{
Overall comparison and component ablation on \bench{}.
LPIPS, VLM-Fid., and Text-CER assess perceptual, holistic, and textual fidelity on standardized renderings.
RFC measures raster fallback on the vector-dominant subset.
Auto Edit-SR evaluates predefined object-level edits, while Rel-SR tests whether attachment, grouping, and containment are preserved after editing.
Editability metrics use each method's native outputs; multi-format scores are averaged across supported formats without per-sample selection.
Human evaluation reports visual fidelity, editing success, and ease.
}

\label{tab:main_results}
\end{table*}

\subsection{Main Results}
\label{sec:main_results}

\paragraph{Fidelity versus native editability.}
Table~\ref{tab:main_results} shows that \method{} achieves the strongest balance between rendered fidelity and native editability.
Existing systems generally exhibit one of two behaviors.
Direct structured generation exposes editable primitives but often sacrifices layout accuracy, textual consistency, or visual detail.
Other systems preserve difficult regions using embedded raster patches, improving local appearance at the cost of native object manipulation.
In contrast, \method{} retains raster representations only for intrinsically image-based content and reconstructs recoverable components using operation-compatible native representations.
Its lower RFC indicates that the fidelity improvement is not obtained through increased raster reuse, while the higher Edit-SR and Rel-SR show that the recovered objects support both direct manipulation and relation-dependent edits.
In particular, attachment, grouping, and containment relations remain valid after the corresponding operations, which is essential for coherent downstream editing.

\paragraph{Practical editing.}
The substantial improvements in automatic Edit-SR and Rel-SR are consistent with the human evaluation, confirming that \method{} provides not only visually plausible outputs but also more practical native editability. Participants can directly replace text, modify shape appearance and geometry, adjust connector styles and endpoints, and manipulate structured chart or table content. These edits can be applied locally while preserving surrounding content and operational relations such as attachment, grouping, and containment. In contrast, rasterized or fragmented representations often lack the required object structure, forcing participants to replace entire regions or manually reconstruct the intended components and their dependencies, resulting in markedly lower task success and editing ease.

\subsection{Qualitative Evaluation}
\label{sec}

Figure~\ref{fig:qualitative_results} presents representative reconstruction and editing results across diverse graphic categories. Existing methods may achieve plausible rendered appearance while leaving text, formulas, connectors, or shapes embedded in raster patches or decomposed into fragmented primitives. Although visually similar, such outputs lose the semantic identity and editable degrees of freedom of the original components, making operations such as modifying text, adjusting shape geometry, recoloring elements, or changing connector attributes difficult to perform natively. In contrast, \method{} preserves the input appearance while recovering these components as independently editable authoring objects. The examples show that localized edits can be applied directly without reconstructing unrelated regions, while surrounding layout and object relations remain coherent.

\subsection{Ablation Study}
\label{sec}

The bottom rows of Table~\ref{tab:main_results} ablate two key components of \method{}.
Removing \textbf{representation selection} replaces type-specific decisions with generic recovery, increasing raster fallback and reducing Edit-SR and Rel-SR.
This confirms that visual similarity alone does not guarantee operation-compatible objects or relation-preserving edits.
Removing \textbf{stage-local verification} disables targeted correction after parsing, recovery, and assembly, degrading fidelity and editability by leaving object, relation, and layout errors unresolved.
Additional refinement-budget and category-specific results are provided in the appendix.

\subsection{Limitations}
\label{sec}

\method{} remains challenged by extremely small text, severe compression, densely overlapping connectors, and highly stylized elements. Ambiguous visual evidence may also cause recognition errors or imperfect representation choices. Because SVG, diagrams.net XML, and PPTX differ in native primitives and relation semantics, some SAR objects or dependencies must be approximated in specific formats. The multi-agent reconstruction and verification process also incurs higher inference cost than one-pass generation. Additional failure cases, format limitations, and efficiency analyses are provided in the appendix.

\section{Conclusion}
\label{sec:conclusion}

We introduced \textbf{raster-to-authoring reconstruction}, which recovers an editing-consistent authoring state from a flattened graphic rather than merely reproducing its appearance. We presented \textbf{\method{}}, a multi-agent framework that reconstructs heterogeneous components in type-appropriate native representations and organizes them with their operational relations in a shared Semantic Authoring Representation. Stage-local verification identifies and repairs localized object, relation, and rendering errors, while deterministic compilers translate the same recovered state into layered SVG, diagrams.net XML, and editable PPTX. We also introduced \textbf{\bench{}} to evaluate fidelity and practical editability. Experiments show that \method{} substantially improves native object- and relation-level editing while reducing avoidable raster reuse and preserving high visual fidelity. More broadly, this work advances graphic reconstruction from pixel reproduction toward recovering structured artifacts that intelligent systems can understand, generate, and revise.

\bibliographystyle{plainnat}
\bibliography{main}

\begin{thebibliography}{44}
\providecommand{\natexlab}[1]{#1}
\providecommand{\url}[1]{\texttt{#1}}
\expandafter\ifx\csname urlstyle\endcsname\relax
  \providecommand{\doi}[1]{doi: #1}\else
  \providecommand{\doi}{doi: \begingroup \urlstyle{rm}\Url}\fi

\bibitem[{BIT-DataLab}(2026)]{EditBanana}
{BIT-DataLab}.
\newblock {Edit Banana}: A framework for converting statistical formats into editable.
\newblock GitHub repository, 2026.
\newblock URL \url{https://github.com/BIT-DataLab/Edit-Banana}.

\bibitem[Carlier et~al.(2020)Carlier, Danelljan, Alahi, and Timofte]{carlier2020deepsvg}
Alexandre Carlier, Martin Danelljan, Alexandre Alahi, and Radu Timofte.
\newblock Deepsvg: A hierarchical generative network for vector graphics animation.
\newblock \emph{Advances in Neural Information Processing Systems}, 33:\penalty0 16351--16361, 2020.

\bibitem[Chen et~al.(2026)Chen, Hong, Mao, Cheng, Liu, Zhang, and Zhang]{chen2026creatiparser}
Weidong Chen, Dexiang Hong, Zhendong Mao, Yutao Cheng, Xinyan Liu, Lei Zhang, and Yongdong Zhang.
\newblock Creatiparser: Generative image parsing of raster graphic designs into editable layers.
\newblock \emph{arXiv preprint arXiv:2604.19632}, 2026.

\bibitem[Cui et~al.(2025)Cui, Yuan, Wang, Li, Du, and Ding]{cui2025draw}
Zhiqing Cui, Jiahao Yuan, Hanqing Wang, Yanshu Li, Chenxu Du, and Zhenglong Ding.
\newblock Draw with thought: Unleashing multimodal reasoning for scientific diagram generation.
\newblock In \emph{Proceedings of the 33rd ACM International Conference on Multimedia}, pages 5050--5059, 2025.

\bibitem[Deng et~al.(2017)Deng, Kanervisto, Ling, and Rush]{deng2017im2markup}
Yuntian Deng, Anssi Kanervisto, Jeffrey Ling, and Alexander~M Rush.
\newblock Image-to-markup generation with coarse-to-fine attention.
\newblock In \emph{International Conference on Machine Learning}, pages 980--989. PMLR, 2017.

\bibitem[Gong et~al.(2025)Gong, Song, Li, Li, and Zhang]{gong2025relationadapter}
Yan Gong, Yiren Song, Yicheng Li, Chenglin Li, and Yin Zhang.
\newblock Relationadapter: Learning and transferring visual relation with diffusion transformers.
\newblock \emph{arXiv preprint arXiv:2506.02528}, 2025.

\bibitem[Hazimeh et~al.(2026)Hazimeh, Wang, Collier, Baechler, Kokiopoulou, and Frossard]{hazimeh2025semantic}
Adam Hazimeh, Ke~Wang, Mark Collier, Gilles Baechler, Efi Kokiopoulou, and Pascal Frossard.
\newblock Semantic document derendering: Svg reconstruction via vision-language modeling.
\newblock In \emph{Proceedings of the AAAI Conference on Artificial Intelligence}, volume~40, pages 4636--4644, 2026.

\bibitem[He et~al.(2026)He, Liu, Memon, Li, Ma, Cho, Ren, Weld, and Krishna]{he2026vfig}
Qijia He, Xunmei Liu, Hammaad Memon, Ziang Li, Zixian Ma, Jaemin Cho, Jason Ren, Daniel~S Weld, and Ranjay Krishna.
\newblock Vfig: Vectorizing complex figures in svg with vision-language models.
\newblock \emph{arXiv preprint arXiv:2603.24575}, 2026.

\bibitem[Hu et~al.(2024)Hu, Yi, Qian, Zhang, Rosin, and Lai]{hu2024supersvg}
Teng Hu, Ran Yi, Baihong Qian, Jiangning Zhang, Paul~L Rosin, and Yu-Kun Lai.
\newblock Supersvg: Superpixel-based scalable vector graphics synthesis.
\newblock In \emph{Proceedings of the IEEE/CVF Conference on Computer Vision and Pattern Recognition}, pages 24892--24901, 2024.

\bibitem[Huang et~al.(2025)Huang, Song, Zhang, Guo, Wang, and Liu]{huang2025arteditor}
Shijie Huang, Yiren Song, Yuxuan Zhang, Hailong Guo, Xueyin Wang, and Jiaming Liu.
\newblock Arteditor: Learning customized instructional image editor from few-shot examples.
\newblock In \emph{Proceedings of the IEEE/CVF International Conference on Computer Vision}, pages 17651--17662, 2025.

\bibitem[Huang et~al.(2026)Huang, Zhou, Gao, Yin, Bai, Liu, Chellappa, Lau, Peng, Nag, and Pramanick]{huang2026scifig}
Siyuan Huang, Yifan Zhou, Yutong Gao, Zi~Yin, Juyang Bai, Xinxin Liu, Rama Chellappa, Chun~Pong Lau, Cheng Peng, Sayan Nag, and Shraman Pramanick.
\newblock Scifig: Towards automating editable figure generation for scientific papers, 2026.
\newblock URL \url{https://arxiv.org/abs/2601.04390}.

\bibitem[Jain et~al.(2023)Jain, Xie, and Abbeel]{jain2023vectorfusion}
Ajay Jain, Amber Xie, and Pieter Abbeel.
\newblock Vectorfusion: Text-to-svg by abstracting pixel-based diffusion models.
\newblock In \emph{Proceedings of the IEEE/CVF Conference on Computer Vision and Pattern Recognition}, pages 1911--1920, 2023.

\bibitem[Li et~al.(2020)Li, Luk{\'a}{\v{c}}, Gharbi, and Ragan-Kelley]{li2020diffvg}
Tzu-Mao Li, Michal Luk{\'a}{\v{c}}, Micha{\"e}l Gharbi, and Jonathan Ragan-Kelley.
\newblock Differentiable vector graphics rasterization for editing and learning.
\newblock \emph{ACM Transactions on Graphics (TOG)}, 39\penalty0 (6):\penalty0 1--15, 2020.

\bibitem[Lin et~al.(2026)Lin, Xie, Zhu, Li, Sun, Gu, Ding, Sun, Guo, Lu, et~al.]{lin2026autofigure}
Zhen Lin, Qiujie Xie, Minjun Zhu, Shichen Li, Qiyao Sun, Enhao Gu, Yiran Ding, Ke~Sun, Fang Guo, Panzhong Lu, et~al.
\newblock Autofigure-edit: Generating editable scientific illustrations via reference-guided styling.
\newblock In \emph{Proceedings of the 64th Annual Meeting of the Association for Computational Linguistics (Volume 3: System Demonstrations)}, pages 57--67, 2026.

\bibitem[Liu et~al.(2023)Liu, Eisenschlos, Piccinno, Krichene, Pang, Lee, Joshi, Chen, Collier, and Altun]{liu2023deplot}
Fangyu Liu, Julian Eisenschlos, Francesco Piccinno, Syrine Krichene, Chenxi Pang, Kenton Lee, Mandar Joshi, Wenhu Chen, Nigel Collier, and Yasemin Altun.
\newblock Deplot: One-shot visual language reasoning by plot-to-table translation.
\newblock In \emph{Findings of the Association for Computational Linguistics: ACL 2023}, pages 10381--10399, 2023.

\bibitem[Lu et~al.(2025)Lu, Zhang, Liu, Wang, and Song]{lu2025easytext}
Runnan Lu, Yuxuan Zhang, Jiaming Liu, Haofan Wang, and Yiren Song.
\newblock Easytext: Controllable diffusion transformer for multilingual text rendering.
\newblock \emph{arXiv preprint arXiv:2505.24417}, 2025.

\bibitem[Luo et~al.(2021)Luo, Li, Wang, and Lin]{luo2021chartocr}
Junyu Luo, Zekun Li, Jinpeng Wang, and Chin-Yew Lin.
\newblock Chartocr: Data extraction from charts images via a deep hybrid framework.
\newblock In \emph{Proceedings of the IEEE/CVF winter conference on applications of computer vision}, pages 1917--1925, 2021.

\bibitem[Ma et~al.(2022)Ma, Zhou, Xu, Sun, Filev, Orlov, Fu, and Shi]{ma2022live}
Xu~Ma, Yuqian Zhou, Xingqian Xu, Bin Sun, Valerii Filev, Nikita Orlov, Yun Fu, and Humphrey Shi.
\newblock Towards layer-wise image vectorization.
\newblock In \emph{Proceedings of the IEEE/CVF Conference on Computer Vision and Pattern Recognition}, pages 16314--16323, 2022.

\bibitem[Mao et~al.(2026{\natexlab{a}})Mao, Wang, Tang, Lin, Li, Yang, Wang, Li, and Tan]{mao2026residual}
Dongxing Mao, Jinpeng Wang, Jiahao Tang, Kevin~Qinghong Lin, Linjie Li, Zhengyuan Yang, Lijuan Wang, Min Li, and Jingru Tan.
\newblock Residual decoder adapter: Id-preserving tokenizer adaption for autoregressive text rendering.
\newblock \emph{arXiv preprint arXiv:2606.01911}, 2026{\natexlab{a}}.

\bibitem[Mao et~al.(2026{\natexlab{b}})Mao, Wang, Li, Yang, and Wang]{mao2026textground4m}
Dongxing Mao, Yilin Wang, Linjie Li, Zhengyuan Yang, and Alex~Jinpeng Wang.
\newblock Textground4m: A prompt-aligned dataset for layout-aware text rendering.
\newblock In \emph{Proceedings of the AAAI Conference on Artificial Intelligence}, volume~40, pages 7918--7926, 2026{\natexlab{b}}.

\bibitem[Reddy et~al.(2021)Reddy, Gharbi, Lukac, and Mitra]{reddy2021im2vec}
Pradyumna Reddy, Michael Gharbi, Michal Lukac, and Niloy~J Mitra.
\newblock Im2vec: Synthesizing vector graphics without vector supervision.
\newblock In \emph{Proceedings of the IEEE/CVF conference on computer vision and pattern recognition}, pages 7342--7351, 2021.

\bibitem[Rodriguez et~al.(2025)Rodriguez, Puri, Agarwal, Laradji, Rodriguez, Rajeswar, Vazquez, Pal, and Pedersoli]{rodriguez2025starvector}
Juan~A Rodriguez, Abhay Puri, Shubham Agarwal, Issam~H Laradji, Pau Rodriguez, Sai Rajeswar, David Vazquez, Christopher Pal, and Marco Pedersoli.
\newblock Starvector: Generating scalable vector graphics code from images and text.
\newblock In \emph{Proceedings of the Computer Vision and Pattern Recognition Conference}, pages 16175--16186, 2025.

\bibitem[Selinger(2003)]{selinger2003potrace}
Peter Selinger.
\newblock Potrace: A polygon-based tracing algorithm, 2003.

\bibitem[Shao et~al.(2026)Shao, Liu, Xu, and Li]{shao2026livefigure}
Chenyang Shao, Jiahe Liu, Fengli Xu, and Yong Li.
\newblock Livefigure: Generating editable scientific illustration with vlm agents.
\newblock \emph{arXiv preprint arXiv:2605.23527}, 2026.

\bibitem[Shi et~al.(2024)Shi, Song, Zhang, Liu, and Zou]{shi2024fonts}
Wenda Shi, Yiren Song, Dengming Zhang, Jiaming Liu, and Xingxing Zou.
\newblock Fonts: Text rendering with typography and style controls.
\newblock \emph{arXiv preprint arXiv:2412.00136}, 2024.

\bibitem[Shi et~al.(2025)Shi, Song, Rao, Zhang, Liu, and Zou]{shi2025wordcon}
Wenda Shi, Yiren Song, Zihan Rao, Dengming Zhang, Jiaming Liu, and Xingxing Zou.
\newblock Wordcon: Word-level typography control in scene text rendering.
\newblock \emph{arXiv preprint arXiv:2506.21276}, 2025.

\bibitem[Song(2022)]{song2022cliptexture}
Yiren Song.
\newblock Cliptexture: Text-driven texture synthesis.
\newblock In \emph{Proceedings of the 30th ACM International Conference on Multimedia}, pages 5468--5476, 2022.

\bibitem[Song and Zhang(2022)]{song2022clipfont}
Yiren Song and Yuxuan Zhang.
\newblock Clipfont: Text guided vector wordart generation.
\newblock In \emph{British Machine Vision Conference}, page 543, 2022.

\bibitem[Song et~al.(2023)Song, Shao, Chen, Zhang, Jing, and Li]{song2023clipvg}
Yiren Song, Xuning Shao, Kang Chen, Weidong Zhang, Zhongliang Jing, and Minzhe Li.
\newblock Clipvg: Text-guided image manipulation using differentiable vector graphics.
\newblock In \emph{Proceedings of the AAAI Conference on Artificial Intelligence}, volume~37, pages 2312--2320, 2023.

\bibitem[Song et~al.(2024)Song, Liu, and Shou]{song2024diffsim}
Yiren Song, Xiaokang Liu, and Mike~Zheng Shou.
\newblock Diffsim: Taming diffusion models for evaluating visual similarity.
\newblock \emph{arXiv preprint arXiv:2412.14580}, 2024.

\bibitem[Song et~al.(2025{\natexlab{a}})Song, Chen, and Shou]{song2025layertracer}
Yiren Song, Danze Chen, and Mike~Zheng Shou.
\newblock Layertracer: Cognitive-aligned layered svg synthesis via diffusion transformer.
\newblock \emph{arXiv preprint arXiv:2502.01105}, 2025{\natexlab{a}}.

\bibitem[Song et~al.(2025{\natexlab{b}})Song, Liu, and Shou]{song2025makeanything}
Yiren Song, Cheng Liu, and Mike~Zheng Shou.
\newblock Makeanything: Harnessing diffusion transformers for multi-domain procedural sequence generation.
\newblock \emph{arXiv preprint arXiv:2502.01572}, 2025{\natexlab{b}}.

\bibitem[Song et~al.(2025{\natexlab{c}})Song, Liu, and Shou]{song2025omniconsistency}
Yiren Song, Cheng Liu, and Mike~Zheng Shou.
\newblock Omniconsistency: Learning style-agnostic consistency from paired stylization data.
\newblock \emph{arXiv preprint arXiv:2505.18445}, 2025{\natexlab{c}}.

\bibitem[Suzuki et~al.(2025)Suzuki, Liu, Inoue, and Yamaguchi]{suzuki2025layerd}
Tomoyuki Suzuki, Kang-Jun Liu, Naoto Inoue, and Kota Yamaguchi.
\newblock Layerd: Decomposing raster graphic designs into layers.
\newblock In \emph{Proceedings of the IEEE/CVF International Conference on Computer Vision}, pages 17783--17792, 2025.

\bibitem[Tang et~al.(2026)Tang, Zhao, Wu, Zhang, Tao, Mao, Wan, Tan, Zeng, Li, and Wang]{tang2026chartscodehierarchicalbenchmark}
Jiahao Tang, Henry~Hengyuan Zhao, Lijian Wu, Zijian Zhang, Yifei Tao, Dongxing Mao, Yang Wan, Jingru Tan, Min Zeng, Min Li, and Alex~Jinpeng Wang.
\newblock From charts to code: A hierarchical benchmark for multimodal models, 2026.
\newblock URL \url{https://arxiv.org/abs/2510.17932}.

\bibitem[Wan et~al.(2024)Wan, Luo, Cai, Song, Zhao, Bai, He, and Gong]{wan2024grid}
Cong Wan, Xiangyang Luo, Zijian Cai, Yiren Song, Yunlong Zhao, Yifan Bai, Yuhang He, and Yihong Gong.
\newblock Grid: Visual layout generation.
\newblock \emph{arXiv preprint arXiv:2412.10718}, 2024.

\bibitem[Wang et~al.(2025)Wang, Zhao, Zhou, Lu, Li, and Song]{wang2025diffdecompose}
Zitong Wang, Hang Zhao, Qianyu Zhou, Xuequan Lu, Xiangtai Li, and Yiren Song.
\newblock Diffdecompose: Layer-wise decomposition of alpha-composited images via diffusion transformers.
\newblock \emph{arXiv preprint arXiv:2505.21541}, 2025.

\bibitem[Wu et~al.(2025{\natexlab{a}})Wu, Su, and Liao]{wu2025chat2svg}
Ronghuan Wu, Wanchao Su, and Jing Liao.
\newblock Chat2svg: Vector graphics generation with large language models and image diffusion models.
\newblock In \emph{Proceedings of the Computer Vision and Pattern Recognition Conference}, pages 23690--23700, 2025{\natexlab{a}}.

\bibitem[Wu et~al.(2025{\natexlab{b}})Wu, Xue, Feng, Wang, and Song]{wu2025mcabench}
Zonglin Wu, Yule Xue, Yaoyao Feng, Xiaolong Wang, and Yiren Song.
\newblock Mca-bench: A multimodal benchmark for evaluating captcha robustness against vlm-based attacks.
\newblock \emph{arXiv preprint arXiv:2506.05982}, 2025{\natexlab{b}}.

\bibitem[Xing et~al.(2024)Xing, Zhou, Wang, Zhang, Xu, and Yu]{xing2024svgdreamer}
Ximing Xing, Haitao Zhou, Chuang Wang, Jing Zhang, Dong Xu, and Qian Yu.
\newblock Svgdreamer: Text guided svg generation with diffusion model.
\newblock In \emph{Proceedings of the IEEE/CVF conference on computer vision and pattern recognition}, pages 4546--4555, 2024.

\bibitem[Zhao et~al.(2026)Zhao, Si, Wang, Wang, Chen, Li, Liang, Sun, and Zhang]{zhao2026crafter}
Haozhe Zhao, Shuzheng Si, Zhenhailong Wang, Zheng Wang, Liang Chen, Xiaotong Li, Zhixiang Liang, Maosong Sun, and Minjia Zhang.
\newblock Crafter: A multi-agent harness for editable scientific figure generation from diverse inputs.
\newblock \emph{arXiv preprint arXiv:2605.30611}, 2026.

\bibitem[Zhao et~al.(2025)Zhao, Luo, Shi, Chen, Wang, Liu, and Sun]{zhao2025chartcoder}
Xuanle Zhao, Xianzhen Luo, Qi~Shi, Chi Chen, Shuo Wang, Zhiyuan Liu, and Maosong Sun.
\newblock Chartcoder: Advancing multimodal large language model for chart-to-code generation.
\newblock In \emph{Proceedings of the 63rd Annual Meeting of the Association for Computational Linguistics (Volume 1: Long Papers)}, pages 7333--7348, 2025.

\bibitem[Zheng et~al.(2026)Zheng, He, Hu, Yu, Yan, Yao, Hou, Zeng, and Wang]{zheng2026chartide}
Xiangxi Zheng, Kuang He, Jiayi Hu, Ping Yu, Rui Yan, Yuan Yao, Peng Hou, Anxiang Zeng, and Alex~Jinpeng Wang.
\newblock Chartide: Data-centric chart-to-code generation via tri-perspective tuning and inquiry-driven evolution.
\newblock In \emph{Proceedings of the 64th Annual Meeting of the Association for Computational Linguistics (Volume 1: Long Papers)}, pages 24966--24985, 2026.

\bibitem[Zhu et~al.(2026)Zhu, Meng, Song, Wei, Li, Pfister, and Yoon]{zhu2026paperbanana}
Dawei Zhu, Rui Meng, Yale Song, Xiyu Wei, Sujian Li, Tomas Pfister, and Jinsung Yoon.
\newblock Paperbanana: Automating academic illustration for ai scientists.
\newblock In \emph{Forty-third International Conference on Machine Learning}, 2026.

\end{thebibliography}

\clearpage
\beginappendix
\startcontents[app]
\begingroup
  \renewcommand{\contentsname}{Appendix Contents}
  \section*{\contentsname}
  \printcontents[app]{}{1}{}
\endgroup
\newpage
\section{Visualization Results and Editability Analysis}
\label{app:Visualization}
\begin{figure*}[htbp]
\centering
\includegraphics[width=\textwidth]{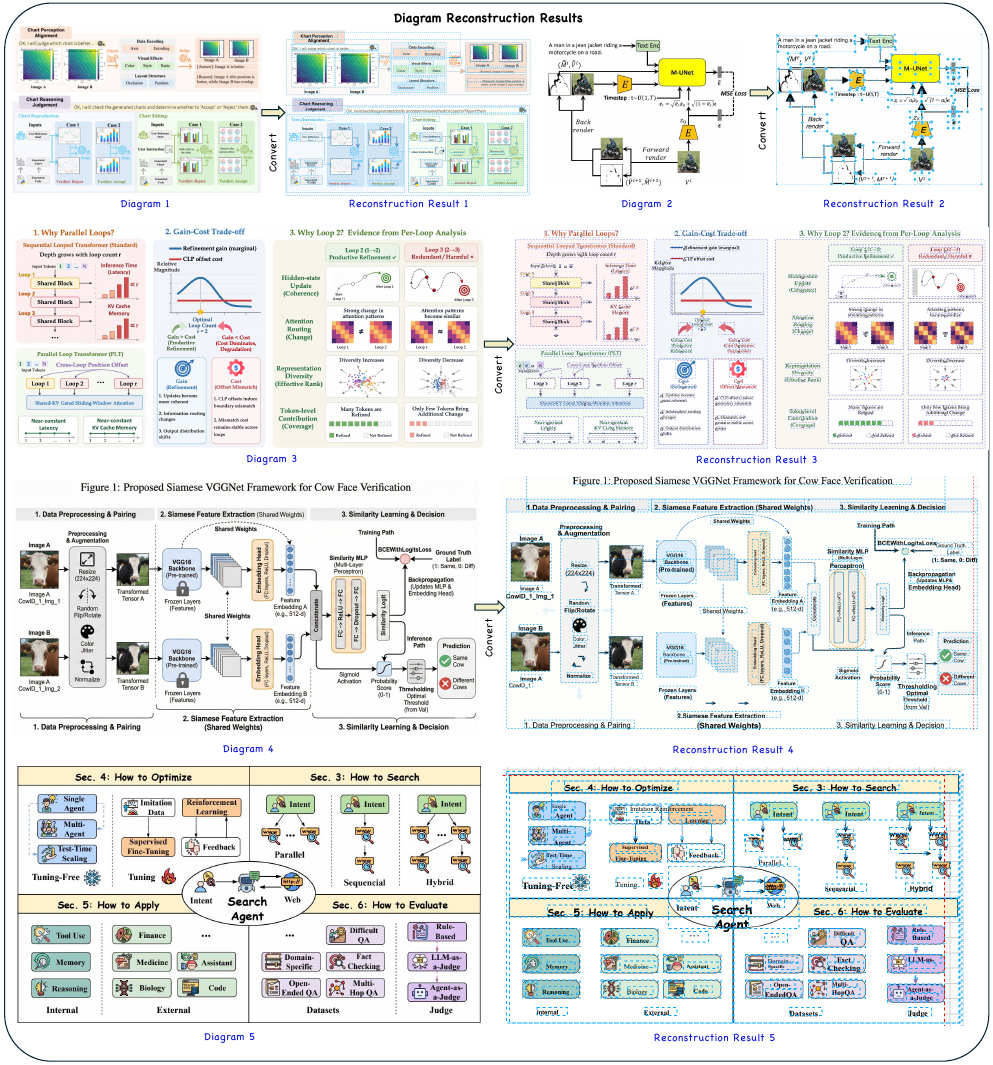}
\caption{
Qualitative reconstruction results for scientific diagrams. The outputs are opened in Inkscape, where blue selection outlines indicate independently editable objects and are not part of the rendered output. \method{} accurately reconstructs diagrams with complex nested layouts and conventional multi-box, multi-arrow structures while maintaining high visual fidelity.
}
\label{fig:supp_diagram1}
\end{figure*}

\begin{figure*}[htbp]
\centering
\includegraphics[width=\textwidth]{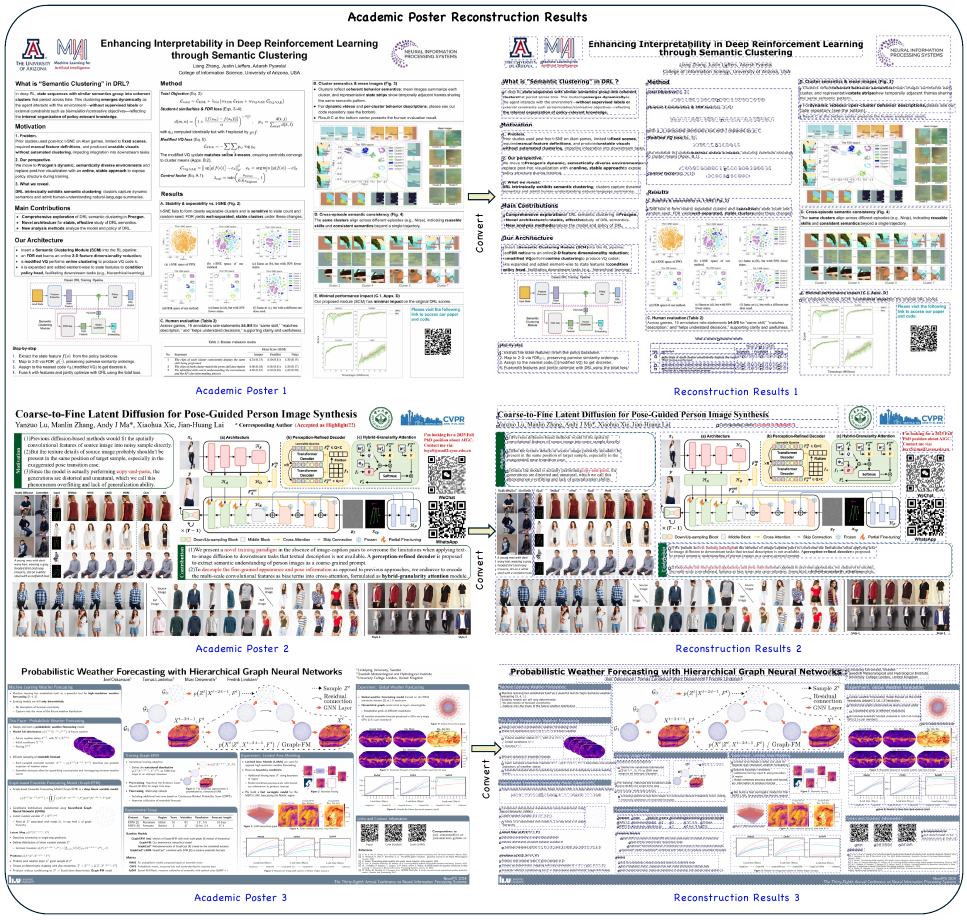}
\caption{
Qualitative reconstruction results for academic posters. \method{} decomposes the complete poster into independently editable text, diagrams, backgrounds, and graphical elements while retaining intrinsically image-based regions as separate raster objects. Blue selection outlines in Inkscape indicate the recovered editable components.
}
\label{fig:supp_aca_poster1}
\end{figure*}

\begin{figure*}[htbp]
\centering
\includegraphics[width=\textwidth]{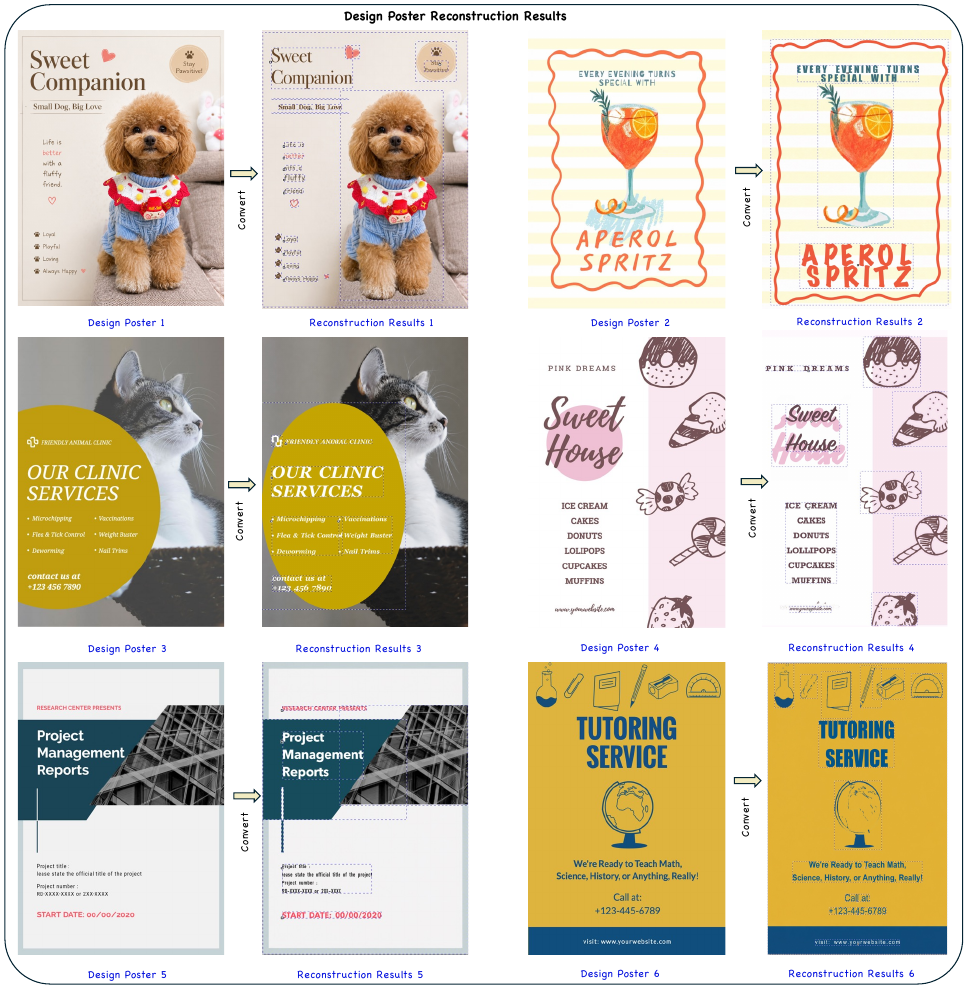}
\caption{
Qualitative reconstruction results for design posters. \method{} separates typography, backgrounds, decorative graphics, and raster image regions into distinct layers, reconstructing recoverable elements as native editable objects and preserving genuine raster content as independent image assets.
}
\label{fig:supp_design_poster1}
\end{figure*}

\begin{figure*}[htbp]

\centering
\includegraphics[width=\textwidth]{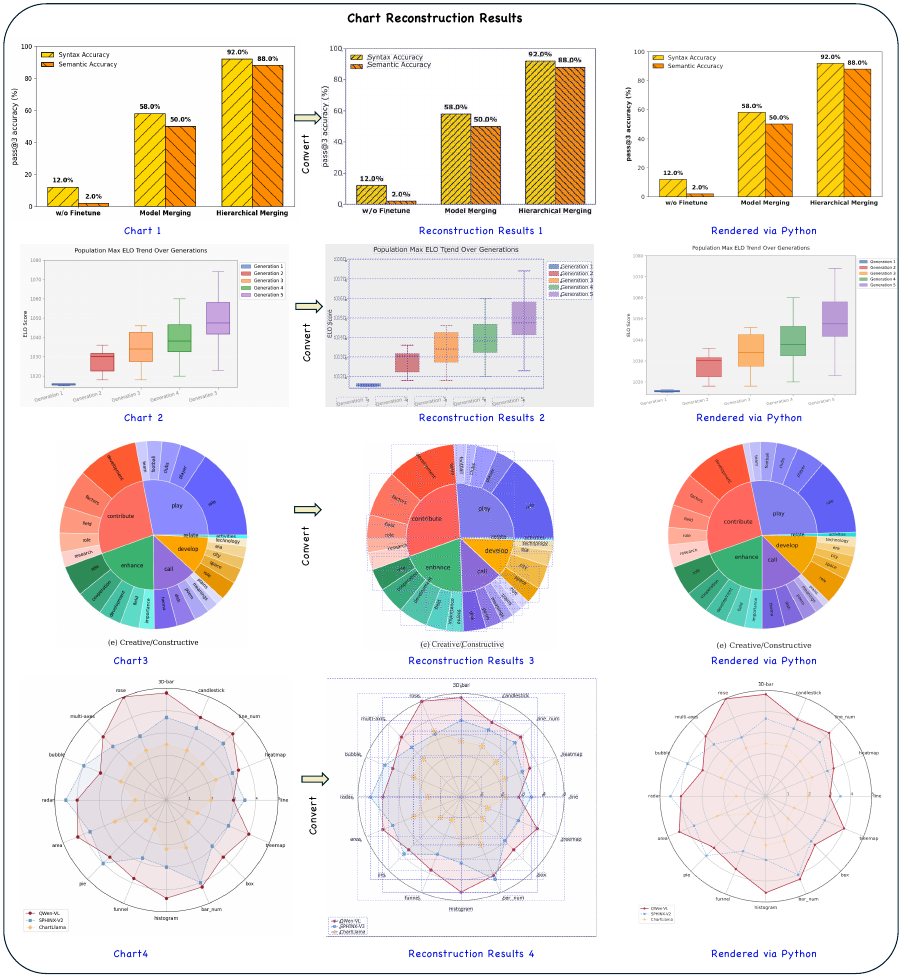}
\caption{
Qualitative reconstruction results for charts. In additio n to editable SVG, diagrams.net-compatible XML, and PPTX artifacts, \method{} recovers Python visualization code when applicable, supporting both object-level modification and data- or program-level editing.
}
\label{fig:supp_chart1}
\end{figure*}

\begin{figure*}[htbp]
\vspace{-4mm}
\centering
\includegraphics[width=\textwidth]{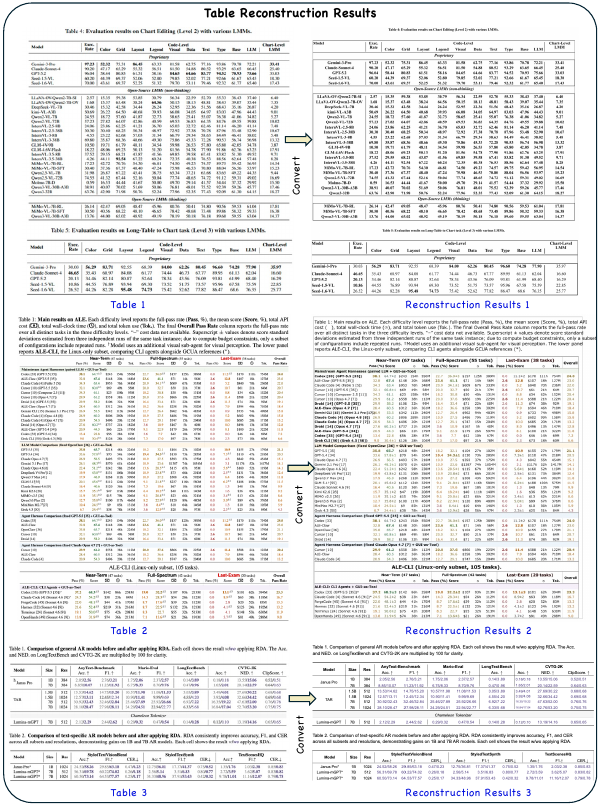}
\caption{
Qualitative reconstruction results for tables. \method{} recovers cell content, row--column organization, borders, fills, typography, and alignment as structured editable content, enabling direct modification of both table values and visual formatting.
}
\label{fig:supp_table1}
\end{figure*}

Figures~\ref{fig:supp_diagram1}--\ref{fig:supp_table1} present additional qualitative results on scientific diagrams, academic posters, design posters, charts, and tables. To explicitly demonstrate the recovered authoring structure, we open the reconstructed layered SVG files in Inkscape. The blue selection outlines indicate independently selectable and editable objects and are interface overlays rather than part of the rendered graphics. Human inspection and VLM-based evaluation show that \method{} preserves the visual appearance, layout, typography, and content of the input while maintaining strong native editability. Manual editing further confirms that only intrinsically image-based regions remain raster objects, whereas recoverable text, shapes, formulas, connectors, charts, tables, and decorative elements are reconstructed in representations appropriate to their editing behavior.

\paragraph{Scientific diagrams.}
Figure~\ref{fig:supp_diagram1} demonstrates the reconstruction of scientific diagrams with substantially different structures. The examples include both visually complex diagrams with nested layouts, composite backgrounds, and heterogeneous components, and conventional network-style diagrams containing multiple boxes and directed arrows. In both cases, \method{} preserves the overall composition while reconstructing text, formulas, shapes, and connectors as independently editable objects. Consequently, users can replace labels, modify shape geometry or appearance, and adjust connector styles without manually recreating the complete diagram. This is particularly valuable for densely connected figures, where reconstructing arrows, boxes, and their spatial organization from scratch would otherwise require considerable manual effort.

\paragraph{Academic posters.}
Figure~\ref{fig:supp_aca_poster1} shows that \method{} can recover the complete authoring structure of an academic poster rather than treating the page as a single flattened image. Academic posters typically combine headings, paragraphs, diagrams, charts, tables, captions, image panels, backgrounds, and decorative elements within a complex multi-column layout. Our agent separates these components and reconstructs recoverable elements in native editable forms while preserving genuine photographs, screenshots, and other intrinsically raster regions as independent image objects. The resulting artifact supports localized modifications to text, diagrams, backgrounds, and page layout without requiring the poster to be recreated from scratch.

\paragraph{Design posters.}
As shown in Figure~\ref{fig:supp_design_poster1}, \method{} performs representation-aware layer decomposition for design posters. Typography, backgrounds, decorative shapes, icons, and raster image regions are separated according to their roles in the original composition. Text and simple graphical elements are converted into native editable objects, while genuine raster content is isolated and, where appropriate, background-removed before being retained as an independent image layer. This reconstruction more closely resembles the layered workflow used in practical graphic design than simple SVG tracing or foreground segmentation. It enables users to replace text, recolor visual elements, reposition image assets, and revise the composition while leaving unrelated regions unchanged.

\paragraph{Charts.}
Figure~\ref{fig:supp_chart1} presents chart reconstruction results. In addition to layered SVG, diagrams.net-compatible XML, and editable PPTX outputs, the chart branch can recover corresponding Python visualization code when the chart structure and underlying data are recoverable. The authoring-format outputs support direct modification of titles, axes, legends, labels, and graphical marks, while the Python code provides an additional interface for editing data values, visual encodings, and rendering parameters. The reconstructed chart can therefore be revised at both the object level and the data or program level, extending editability beyond visual element manipulation alone.

\paragraph{Tables.}
Figure~\ref{fig:supp_table1} illustrates the reconstruction of tables as structured editable content rather than flattened images or collections of disconnected text boxes. \method{} recovers cell values together with row--column organization, borders, fills, typography, and alignment. The recovered structure is exported as native table objects where supported and as organized editable cells in other target formats. Users can therefore directly replace table entries and modify cell backgrounds, border styles, text appearance, and alignment without reconstructing the table layout.

\section{Implementation Details}
\label{app:implementation}

\label{app:implementation}

\subsection{System Overview}

\method{} is a category-routed multi-agent system that reconstructs a flattened raster graphic into a format-agnostic authoring state and compiles that state into editable artifacts. Given an input image $I$, a graphic-level router selects one of five category-specialized branches: scientific diagrams, charts, tables, academic posters, or design posters. Natural-image content is not treated as a separate benchmark category; photographs, screenshots, microscopy panels, and other intrinsically image-based regions are handled by the raster-region recovery operator within the selected branch.

Although the specialized branches use category-dependent parsing and recovery tools, they share the following four-stage workflow:

\begin{enumerate}
\item \textbf{Evidence parsing.}
The selected parser decomposes the input into candidate visual regions and extracts structural, semantic, relational, and visual evidence.

\item \textbf{Representation-aware recovery.}
Type-specific recovery agents reconstruct text, formulas, shapes, connectors, icons, charts, tables, and raster panels using representations that preserve their intended editing operations.

\item \textbf{Relation-aware assembly.}
Recovered objects are consolidated into the Semantic Authoring Representation (SAR), rendering order is resolved, and operational relations are instantiated and checked.

\item \textbf{Deterministic compilation.}
Format-specific compilers translate the verified SAR into layered SVG, diagrams.net-compatible XML, and editable PPTX.
\end{enumerate}

Stage-local verification operates after parsing, object recovery, and relation assembly. It checks object validity, relation consistency, and rendering fidelity, then routes each localized error to the responsible stage while preserving validated SAR entries.

\subsection{Category-Specialized Branches}

The router selects one of the following branches:

\begin{itemize}
\item \textbf{Scientific-diagram branch.}
This branch emphasizes text, formulas, parameterized shapes, connectors, directed relations, grouping, and nested diagram structure.

\item \textbf{Chart branch.}
This branch emphasizes plot-region detection, axes, ticks, legends, labels, visual marks, recoverable data, and encoding structure.

\item \textbf{Table branch.}
This branch emphasizes grid structure, row and column topology, merged cells, headers, cell-level text, borders, and alignment.

\item \textbf{Academic-poster branch.}
This branch emphasizes multi-column layout, section hierarchy, text blocks, formulas, figure panels, charts, tables, captions, and section-level grouping.

\item \textbf{Design-poster branch.}
This branch emphasizes typography, visual hierarchy, decorative shapes, icons, image panels, masks, and layered composition.
\end{itemize}

All branches produce records under the same SAR schema and use the same compiler interfaces. Category routing therefore changes the evidence extraction and recovery operators, but not the semantics of the downstream authoring state.

\subsection{Agent Topology and Structured Interfaces}

Table~\ref{tab:agent_interfaces} summarizes the role and structured output of each agent. Agent outputs are validated against fixed schemas before entering the next stage. Malformed records are rejected and retried under the same predefined retry policy.

\begin{table*}[t]
\centering
\caption{Agent roles and structured interfaces in \method{}.}
\label{tab:agent_interfaces}

\vspace{2pt}

\fontsize{9.5pt}{11.5pt}\selectfont
\setlength{\tabcolsep}{5pt}
\renewcommand{\arraystretch}{1.15}

\begin{tabularx}{\textwidth}{
    >{\raggedright\arraybackslash}p{0.16\textwidth}
    >{\raggedright\arraybackslash}p{0.31\textwidth}
    >{\raggedright\arraybackslash}X
}
\toprule

\textbf{Agent}
& \textbf{Primary Role}
& \textbf{Structured Output}
\\
\midrule

\textbf{Graphic router}
& Select a category-specialized branch
& Category, confidence, and routing evidence
\\

\textbf{Parsing agent}
& Identify candidates and aggregate multi-source evidence
& Regions, provisional types, geometry, evidence, and relation proposals
\\

\textbf{Recovery agents}
& Select representations and recover editable payloads
& SAR object records and recovery confidence
\\

\textbf{Relation assembler}
& Instantiate and resolve cross-object relations
& Typed relation records, ordering information, and conflict-resolution log
\\

\textbf{Stage-local verifier}
& Localize object, relation, and rendering errors
& Responsible stage, affected IDs, error type, evidence, and confidence
\\

\textbf{Format compiler}
& Serialize verified SAR deterministically
& Native artifact and export/approximation log
\\

\bottomrule
\end{tabularx}

\end{table*}

\subsection{Semantic Authoring Representation}

The reconstruction state is the Semantic Authoring Representation defined in the main paper:
\begin{equation}
\mathcal{A}=(O,E),
\label{eq:app_sar}
\end{equation}
where $O=\{o_i\}_{i=1}^{N}$ is the reconstructed object set and $E$ is the relation set. Each object is represented as
\begin{equation}
o_i=(\tau_i,r_i,c_i,b_i,g_i,a_i,z_i),
\label{eq:app_object}
\end{equation}
where $\tau_i$ denotes semantic type, $r_i$ the selected authoring representation, $c_i$ the type-specific editable payload, $b_i$ spatial support, $g_i$ geometry, $a_i$ appearance, and $z_i$ rendering order.

A relation record is represented as
\begin{equation}
e_k=(\rho_k,\mathcal{S}_k,\mathcal{T}_k,\psi_k),
\label{eq:app_relation}
\end{equation}
where $\rho_k$ is the relation type, $\mathcal{S}_k$ and $\mathcal{T}_k$ identify the participating source and target objects, and $\psi_k$ stores relation-specific parameters. For example, an attachment record stores a connector endpoint and its attached object, a containment record stores a container and contained object, and a grouping record stores group membership and the group transform.

The distinction between semantic type $\tau_i$ and authoring representation $r_i$ is central. Semantic type specifies what an object represents, whereas authoring representation determines which editing operations it supports. A formula region may be recognized correctly but still be represented as a symbolic expression, unrelated vector paths, or a raster crop. Likewise, an arrow-shaped mark is not a functional connector unless its shaft, markers, endpoints, and attachment dependencies are represented.

The payload $c_i$ is type-dependent. It may contain character content and typography, symbolic expressions, chart data and encodings, table cells and topology, connector endpoints and markers, parameterized primitive attributes, vector geometry, or a raster asset with crop and mask information. The outer schema remains shared across types, allowing heterogeneous components to coexist without reducing them to a single low-level representation.

\subsection{Evidence Parsing}

The selected parsing branch extracts four complementary evidence types:

\begin{itemize}
\item \textbf{Structural evidence} describes panels, layout, reading order, hierarchy, grouping cues, and repeated organization.
\item \textbf{Semantic evidence} identifies textual, symbolic, graphical, data-bearing, and intrinsically raster content.
\item \textbf{Relational evidence} proposes alignment, association, containment, grouping, hierarchy, and attachment dependencies.
\item \textbf{Visual evidence} records spatial support, geometry, appearance, masks, local rendering cues, and evidence for raster preservation.
\end{itemize}

A candidate record contains its region, provisional semantic type, confidence, neighboring context, proposed relations, and the evidence supporting each decision. This record is passed to representation selection rather than directly serialized into an output format.

\subsection{Representation Selection and Recovery Operators}

For each candidate $p_i$, the system jointly selects a semantic type and an admissible authoring representation:
\begin{equation}
(\tau_i,r_i)=\Phi(p_i,I,\mathcal{C}_i),
\qquad
r_i\in\mathcal{R}(\tau_i),
\label{eq:app_rep_selection}
\end{equation}
where $\mathcal{C}_i$ contains neighboring candidates, structural predictions, relation proposals, and the current SAR state. The selection operator favors the most structured representation reliably supported by the available evidence rather than the representation that most directly reproduces visible pixels. Table~\ref{tab:format_mapping} summarizes the preferred representation, editable payload, and minimum validity conditions of each recovery operator.

\paragraph{Text recovery.}
The text operator recovers character content together with font family, size, weight, style, alignment, line spacing, color, and spatial support. Multi-line regions preserve line grouping, paragraph organization, and reading order. Low-confidence text is explicitly marked for stage-local verification and repeated correction rather than silently converted into vector outlines or raster content.

\paragraph{Formula recovery.}
The formula operator reconstructs each detected expression as an editable symbolic representation. It jointly recovers the symbolic payload, spatial support, typography, and display attributes, while keeping the symbolic content separate from its rendered geometry. When the initial result is invalid or visually inconsistent, the system performs additional recognition and verification rounds until a parseable and editable formula is obtained. Formula regions are therefore not replaced with visual approximations or localized raster crops; unsuccessful recovery after the predefined verification budget is recorded as a reconstruction failure.

\paragraph{Shape recovery.}
The shape operator decomposes a candidate region into its constituent shapes and reconstructs each component as an independently editable object. Whenever possible, shapes are represented using native parameterized primitives with editable geometry, fill, stroke, corner radius, opacity, and transforms. Irregular shapes are recovered as structured vector paths while retaining independent object boundaries and rendering order. This decomposition prevents visually distinct shapes from being merged into a single raster patch or inseparable vector group.

\begin{table*}[t]
\centering
\caption{
Format-specific mappings from SAR to native or structured editable authoring objects.
Chart exports preserve the recovered data and visualization specification, while icon rasterization is used only when faithful editable reconstruction cannot be achieved.
}
\label{tab:format_mapping}

\vspace{2pt}

\fontsize{9.2pt}{11.2pt}\selectfont
\setlength{\tabcolsep}{4.0pt}
\renewcommand{\arraystretch}{1.12}

\begin{tabularx}{\textwidth}{
    >{\raggedright\arraybackslash}p{0.085\textwidth}
    >{\raggedright\arraybackslash}X
    >{\raggedright\arraybackslash}X
    >{\raggedright\arraybackslash}X
}
\toprule

\textbf{SAR Content}
& \textbf{Layered SVG}
& \textbf{diagrams.net XML}
& \textbf{PPTX}
\\

\midrule

\textbf{Text}
& Editable text elements with typography
& Editable text cells with style attributes
& Editable text frames and runs
\\

\textbf{Formula}
& Editable formula group with symbolic payload
& Editable formula group with symbolic payload
& Editable equation or structured formula object
\\

\textbf{Shape}
& Independently editable primitives or vector paths
& Independently styled vertex geometries
& Editable AutoShapes or freeform shapes
\\

\textbf{Icon}
& Editable primitive/group; otherwise masked raster icon
& Editable grouped cells; otherwise background-removed image cell
& Editable shape group; otherwise background-removed picture
\\

\textbf{Connector}
& Native editable connector with endpoint and attachment metadata
& Native edge with source/target references
& Native connector with endpoint attachments
\\

\textbf{Group}
& Group element with shared transform
& Parent--child cell structure
& Native group shape
\\

\textbf{Table}
& Editable cell grid with borders, spans, and text
& Editable cell structure with borders and merged regions
& Native editable table with cell-level content and formatting
\\

\textbf{Chart}
& Editable marks with data, encodings, and code metadata
& Grouped editable chart objects with data and code metadata
& Native editable chart with embedded data and code metadata
\\

\textbf{Raster panel}
& Positioned image element with crop, mask, and transform
& Independently editable image cell
& Independently editable picture shape
\\

\bottomrule
\end{tabularx}

\end{table*}

\paragraph{Icon recovery.}
Icons are first reconstructed by a vision-language model as editable native primitives or structured vector groups. The recovered icon is rendered and compared with the corresponding input region through stage-local verification. When the reconstructed icon sufficiently preserves the original geometry and appearance, the editable representation is retained. When faithful editable reconstruction cannot be obtained, the system instead extracts the icon from the input, removes its background, and preserves it as an independently positioned raster icon. This policy prioritizes native editability while avoiding visually inaccurate vector approximations.

\paragraph{Connector recovery.}
The connector operator uses a vision-language model to interpret visible lines, arrows, and their surrounding objects, and reconstructs them as native functional connectors rather than independent line segments or rasterized arrows. The recovered payload includes shaft geometry, routing style, endpoint positions, arrowheads or other markers, and source--target attachment proposals. A connector is accepted only when its native representation, endpoint geometry, and corresponding attachment records are valid.

\paragraph{Chart recovery.}
The chart operator reconstructs each chart as a fully editable chart representation, including plot structure, underlying data, axes, ticks, labels, legends, visual marks, and encoding attributes. In addition to the editable chart object, the system generates the corresponding Python visualization code that reproduces the recovered chart from its structured data and visual configuration. The code, data, and chart-specific attributes are stored together in the editable payload, enabling both direct authoring-tool manipulation and programmatic regeneration or modification.

\paragraph{Table recovery.}
The table operator reconstructs each detected table as a fully editable table object. It recovers row and column topology, individual cells, merged-cell spans, headers, cell text and numerical values, borders, line widths, fills, alignment, typography, and spatial layout. Both the table structure and its visible content remain editable at the cell level, allowing users to modify numbers, labels, formatting, and table lines without replacing or redrawing the complete table.

\paragraph{Raster-panel recovery.}
Photographs, microscopy images, screenshots, continuous-tone illustrations, and other intrinsically image-based regions are retained as independently positioned and masked raster objects. Their crop, mask, opacity, geometry, and rendering order remain editable. Such raster panels are not treated as reconstruction failures when raster representation is semantically appropriate to their content.

\subsection{Relation-Aware Assembly}

The relation assembler resolves rendering order and instantiates relations in $E$ from structural evidence, agent proposals, and type-specific constraints. It verifies that every relation refers to existing objects, that participating types are compatible, and that conflicting proposals are resolved before entering SAR.

The following relation types are represented:

\begin{itemize}
\item \textbf{Grouping:} objects are manipulated under a shared group transform.
\item \textbf{Containment:} an object belongs to a container or panel.
\item \textbf{Association:} labels, legends, or annotations refer to another object.
\item \textbf{Alignment:} objects participate in a shared alignment constraint.
\item \textbf{Hierarchy:} objects form parent--child or nested organizational structure.
\item \textbf{Attachment:} a connector endpoint is attached to a source or target object.
\end{itemize}

Attachment, grouping, and containment are evaluated through executable relation-preserving edits. Association, alignment, and hierarchy remain represented in SAR and are inspected through structural validation and qualitative analysis.

\subsection{Stage-Local Verification and Repair}

At refinement step $t$, the current authoring state is $\mathcal{A}^{(t)}$, and its standardized rendering is
\begin{equation}
R^{(t)}=\mathcal{R}\!\left(\mathcal{A}^{(t)}\right).
\label{eq:app_render_state}
\end{equation}
A stage-local verifier returns a structured error record
\begin{equation}
e^{(t)}=
\left(
s^{(t)},S^{(t)},\kappa^{(t)},\ell^{(t)},q^{(t)}
\right),
\label{eq:app_error_record}
\end{equation}
where $s^{(t)}$ is the responsible stage, $S^{(t)}$ the affected SAR subset, $\kappa^{(t)}$ the error type, $\ell^{(t)}$ localized visual or structural evidence, and $q^{(t)}$ verifier confidence.

Verification covers three aspects:

\begin{itemize}
\item \textbf{Object validity} checks compatibility among semantic type, representation, payload, geometry, appearance, and supported editable degrees of freedom.
\item \textbf{Relation consistency} checks existence, type compatibility, and geometric consistency of grouping, containment, association, hierarchy, alignment, and attachment records.
\item \textbf{Rendering fidelity} checks localized discrepancies in visible content, layout, typography, geometry, appearance, and composition.
\end{itemize}

Only the affected subset is revised:
\begin{equation}
\mathcal{A}^{(t+1)}_{S^{(t)}}=
U_{s^{(t)}}\!\left(I,\mathcal{A}^{(t)},e^{(t)}\right),
\qquad
\mathcal{A}^{(t+1)}_{\overline{S}^{(t)}}=
\mathcal{A}^{(t)}_{\overline{S}^{(t)}}.
\label{eq:app_localized_repair}
\end{equation}
Refinement terminates when no actionable error above the fixed confidence threshold remains or the maximum budget $T_{\max}$ is reached. Validated objects and relations are retained rather than regenerated.

\subsection{Deterministic Multi-Format Compilation}

For target format $f$, compiler $C_f$ maps SAR objects, payloads, relations, and rendering order into the primitives supported by that environment. The compiler does not invoke a generative model and may serialize only information already present in SAR. For a fixed SAR and backend configuration, compilation is deterministic.

Table~\ref{tab:format_mapping} summarizes the intended mappings. When a backend lacks a direct counterpart, the compiler emits a structured composition and records the approximation in an export log.

When a method supports multiple native output formats, each format is evaluated independently. For metric $q$ and method $m$, the reported multi-format score is
\begin{equation}
q_m=
\frac{1}{|\mathcal{F}_m|}
\sum_{f\in\mathcal{F}_m}q_{m,f},
\label{eq:format_average}
\end{equation}
where $\mathcal{F}_m$ is the set of officially supported outputs. No per-sample best-format selection is performed. Per-format results for \method{} are additionally reported in Sec.~\ref{app:multiformat}.

\subsection{Output Validation}

Every compiled artifact undergoes format-specific validation:

\begin{itemize}
\item Layered SVG must be valid XML, contain a valid canvas, expose at least one visible element, and render successfully.
\item diagrams.net-compatible XML must be parseable, contain a valid graph model, open in the target environment, and render successfully.
\item PPTX must be loadable by an Office Open XML parser, contain a valid slide canvas, and render without corruption.
\end{itemize}

Validation checks artifact integrity only; it does not repair or reinterpret the generated content.

\subsection{Implementation Configuration}
Table~\ref{tab:implementation_details} summarizes the implementation details of \method{}, including the models, software libraries, reconstruction modules, verification settings, and export backends. The graphic router and most parsing, recovery, and verification components are built on GPT-5.5, while PaddleOCR and GPT-image-2 are used for text recognition and image-oriented component recovery, respectively. Type-specific modules handle formulas, shapes, connectors, charts, and tables, and deterministic compilers export the recovered authoring state to layered SVG, diagrams.net XML, and editable PPTX. Unless otherwise specified, all model snapshots, prompts, decoding parameters, retry budgets, validation thresholds, and rendering settings are fixed across all benchmark samples to ensure consistent evaluation.

\begin{table*}[t]
\centering
\caption{
Implementation details of \method{}.
All model snapshots, prompts, decoding parameters, software versions, and
verification thresholds are fixed across benchmark samples.
}
\label{tab:implementation_details}

\vspace{2pt}

\footnotesize
\setlength{\tabcolsep}{4pt}
\renewcommand{\arraystretch}{1.15}

\begin{tabularx}{\textwidth}{
    >{\raggedright\arraybackslash}p{0.12\textwidth}
    >{\raggedright\arraybackslash}p{0.19\textwidth}
    >{\raggedright\arraybackslash}p{0.24\textwidth}
    >{\raggedright\arraybackslash}X
}
\toprule
\textbf{Component}
& \textbf{Role}
& \textbf{Implementation / Version}
& \textbf{Key Configuration}
\\
\midrule

Graphic router
& Five-category branch selection
& GPT-5.5 (API, July 2026 snapshot)
& Prompt v1.0; temperature 0; JSON output; deterministic top-1 routing.
\\

Backbone MLLM
& Parsing, representation selection, and recovery
& GPT-5.5 (API, July 2026 snapshot)
& Longest image side 2048 px; max 16,384 output tokens; temperature 0.2; max 3 retries.
\\

Text/OCR recovery
& Text detection, recognition, and typography recovery
& PaddleOCR 3.7.0 with PP-OCRv5 server models
& English/Chinese recognition; confidence threshold 0.60; orientation and unwarping disabled.
\\

Formula recovery
& Iterative editable symbolic reconstruction
& GPT-5.5 with a custom formula-validation module
& LaTeX output; max 5 attempts; syntax parsing and render validation; no raster fallback.
\\

\addlinespace[2pt]

Shape recovery
& Decomposition into independently editable shapes
& GPT-5.5 with a custom Python/SVG geometry module
& Native primitives preferred; vector paths for irregular shapes; 2-px geometry tolerance; max 3 retries.
\\

Icon recovery
& Editable icon reconstruction or isolated raster preservation
& GPT-image-2 with GPT-5.5 visual verification
& Max 3 reconstruction attempts; acceptance score $\geq 4/5$; otherwise background-removed raster icon.
\\

Connector recovery
& Native connector and endpoint-attachment reconstruction
& GPT-5.5 with a custom connector-construction module
& Straight, elbow, and curved routing; 8-px endpoint tolerance; native marker library; max 3 retries.
\\

Chart recovery
& Editable chart, structured data, encodings, and Python code
& GPT-5.5; Python 3.11; Matplotlib 3.11.1
& Bar, line, scatter, area, pie, heatmap, and box plots; 30-s execution timeout; max 5 repair attempts.
\\

Table recovery
& Editable cells, topology, values, and table borders
& GPT-5.5 with PaddleOCR 3.7.0
& Cell-level reconstruction; merged-cell support; 3-px grid tolerance; editable text, values, fills, and borders.
\\

\addlinespace[2pt]

Stage-local verifier
& Localized object, relation, and rendering repair
& GPT-5.5 with CairoSVG 2.7.1 rendering
& $T_{\max}=3$; temperature 0; confidence threshold 0.75; targeted repair without regenerating validated objects.
\\

\addlinespace[2pt]

SVG backend
& Layered SVG compilation, validation, and rendering
& Custom Python SVG compiler; CairoSVG 2.7.1
& SVG 1.1-compatible output; XML validation; Noto/DejaVu font fallback; standardized rendering at 144 DPI.
\\

diagrams.net backend
& Editable XML compilation, validation, and rendering
& Custom mxGraph XML compiler; diagrams.net Desktop 31.1.5
& \texttt{mxGraphModel} schema; native vertices, groups, and edges; source--target endpoint references.
\\

PPTX backend
& Editable presentation compilation and validation
& Custom OOXML compiler; python-pptx 1.0.0
& Native text frames, AutoShapes, connectors, tables, charts, groups, and picture objects.
\\

\bottomrule
\end{tabularx}

\end{table*}

\subsection{Inference Settings and Reproducibility}

Unless otherwise stated, inputs are resized while preserving aspect ratio and evaluated on a fixed target canvas of $[W\times H]$. The original aspect ratio is retained during reconstruction. Deterministic decoding is used whenever supported. For stochastic components, fixed seeds and the complete decoding configuration are reported.

For every sample, we record wall-clock inference time, input and output tokens, external tool calls, number of verification rounds, compiler time, artifact validity, and API cost. No manual correction, manual layer cleanup, benchmark-specific prompt modification, or post-hoc artifact repair is applied before evaluation.

Prompts, structured schemas, model and software versions, parser settings, rendering commands, format-specific compilers, and evaluation scripts will accompany the released evaluation package.

\section{\bench{}}
\label{app:benchmark}

\subsection{Benchmark Scope}

\bench{} contains 250 flattened raster graphics across the same five categories used in the main paper:

\begin{enumerate}
\item scientific diagrams;
\item charts;
\item tables;
\item academic posters;
\item design posters.
\end{enumerate}

The benchmark covers text, formulas, shapes, connectors, icons, chart elements, table structures, and raster panels, as shown in Table~\ref{tab:benchmark_statistics}. It evaluates realistic settings in which only the flattened image is available. Original editable source files are not provided to evaluated methods and are not required by the evaluation protocol.

\begin{table}[t]
\centering
\caption{Category distribution of \bench{}.}
\label{tab:benchmark_statistics}

\vspace{1pt}

\fontsize{8.5pt}{10pt}\selectfont
\setlength{\tabcolsep}{3.5pt}
\renewcommand{\arraystretch}{1.08}

\begin{tabularx}{0.96\columnwidth}{
    >{\raggedright\arraybackslash}p{0.27\columnwidth}
    >{\centering\arraybackslash}p{0.13\columnwidth}
    >{\raggedright\arraybackslash}X
}
\toprule
\textbf{Category}
& \textbf{\# Samples}
& \textbf{Typical Components}
\\
\midrule

Scientific diagrams
& 100
& Text, formula, shape, connector
\\

Charts
& 25
& Axes, marks, legends, labels
\\

Tables
& 25
& Cells, headers, spans, borders
\\

Academic posters
& 50
& Sections, figures, tables, formulas
\\

Design posters
& 50
& Typography, icons, decorative shapes
\\

\midrule

\textbf{Total}
& \textbf{250}
& --
\\

\bottomrule
\end{tabularx}

\end{table}

\subsection{Benchmark Collection and Candidate Sources}
\label{app:data_collection}

\bench{} was constructed from category-specific candidate pools collected from heterogeneous sources. For scientific diagrams, five undergraduate annotators in computer science collected approximately 500 figures from papers published in 2025--2026 across multiple STEM disciplines. These candidates covered conventional multi-box and multi-arrow diagrams as well as more complex figures containing nested structures, composite backgrounds, formulas, icons, and heterogeneous visual elements.

For academic posters, the annotators collected approximately 150 posters from publicly accessible poster pages and galleries associated with major conferences, including NeurIPS, ICML, and CVPR. For design posters, approximately 150 candidates were collected from publicly accessible examples on design platforms such as Canva and Lovart. A further candidate pool of approximately 150 chart and table images was assembled and screened using the same procedure. The final benchmark contains 100 scientific diagrams, 50 academic posters, 50 design posters, 25 charts, and 25 tables, for a total of 250 samples.

\subsection{Human Screening and Final Selection}
\label{app:data_selection}

All five annotators independently reviewed the collected candidates before the final benchmark was constructed. The screening criteria considered visual clarity, compositional quality, structural complexity, element diversity, editing relevance, source resolution, and reconstruction difficulty. For example, diagram selection considered the number and diversity of objects, connector density, nesting depth, background complexity, and the presence of formulas or raster regions. Poster selection additionally considered layout organization, typography, visual hierarchy, and the diversity of embedded content.

The selection process was not based solely on visual attractiveness. We also retained samples covering different structural patterns and a broad range of reconstruction difficulty. After independent screening, the annotators aggregated their selections and discussed inconsistent judgments until reaching a consensus. The final subset was chosen to balance visual quality, category coverage, structural diversity, and human-rated difficulty, while avoiding excessive concentration on any particular source, layout template, or visual style.




\subsection{Duplicate and Near-Duplicate Removal}
\label{app:duplicate_removal}

Duplicate removal is performed on the candidate pools before final benchmark selection. Candidate pairs are first proposed using perceptual image hashes and visual embeddings. This procedure identifies not only exact duplicates but also near-duplicate variants produced by resizing, cropping, compression, or minor visual modification. Candidate pairs above fixed similarity thresholds are then manually reviewed.

When two candidates contain substantially the same visual content, only one representative is retained, with preference given to the version having higher resolution, more complete content, and clearer source provenance. Near-duplicate templates with only minor textual or stylistic changes are also removed when they do not introduce meaningful differences in reconstruction structure or difficulty. The embedding model, hash algorithm, similarity thresholds, number of proposed pairs, and number of removed samples are documented with the benchmark release.

\subsection{Difficulty Annotation}

The human-rated difficulty distribution shown in the main paper is obtained from the raster inputs alone, without exposing method outputs. Each sample is independently rated by 5 annotators using the following five-point reconstruction-difficulty scale:

\begin{itemize}
\item 1: few large elements and simple layout;
\item 2: limited heterogeneity with clear separation;
\item 3: moderate element density or mixed content types;
\item 4: dense layout, small text, complex relations, or multiple panels;
\item 5: severe density, overlap, stylization, degradation, or highly heterogeneous content.
\end{itemize}

The sample difficulty is computed as the mean rating across five annotators. Inter-rater reliability is assessed using a two-way random-effects, absolute-agreement, average-measures intraclass correlation coefficient, ICC(2,5), with a 95\% confidence interval.

\subsection{Evaluation Subsets}

Different metrics are computed on subsets determined by their applicability.

\paragraph{Full fidelity set.}
All 250 images are used for rendered fidelity, textual consistency, and output-validity evaluation. Qualitative examples are selected from this set using the protocol described in previous section.

\paragraph{Vector-dominant set.}
RFC is primarily evaluated on scientific diagrams, charts, and tables, whose visible foreground content is expected to be reconstructable predominantly as native text, vector, connector, chart, or table objects. Raster occupancy for academic and design posters is reported separately because such graphics may contain semantically appropriate raster panels.

\paragraph{Executable-edit set.}
For each object-level editing operation, a source-image detector identifies applicable targets. Only targets whose confidence exceeds the fixed operation-specific threshold are included in that operation's denominator.

\paragraph{Relation-edit set.}
Rel-SR is evaluated on source images containing verified attachment, grouping, or containment relations. The set uses sparse operation-level relation records rather than exhaustive object-graph annotations.

\paragraph{Human-study set.}
A category-stratified subset is selected before method evaluation. Each sample is paired with a predefined object-level or relation-dependent editing task drawn from the automatic operation suite.

\paragraph{Common-domain set.}
To separate category coverage from within-domain performance, we additionally evaluate all methods on the intersection of their officially supported categories. This subset is fixed before evaluation and shared across all common-domain metrics.

\subsection{Development Protocol and Leakage Control}

Prompts, detector thresholds, matching thresholds, edit definitions, verifier policies, and retry budgets are fixed before final benchmark evaluation. Development uses a disjoint set excluded from \bench{}. Benchmark images and their near-duplicates are excluded from prompt demonstrations and manually curated few-shot examples. No per-sample prompt modification, manual correction, or metric-guided output selection is permitted.

\section{Baselines and Fair Comparison}
\label{app:baselines}

\subsection{Compared Methods}

The primary comparison includes the following task-aligned raster-to-editable systems:

\begin{itemize}
\item \textbf{Edit-Banana}, evaluated in its officially supported diagrams.net-compatible XML representation;
\item \textbf{AutoFigure-Edit}, evaluated in its officially supported SVG representation;
\item \textbf{CraftEditor}, evaluated in its officially supported SVG representation;
\item \textbf{Direct MLLM}, implemented with GPT-5.5 and a fixed direct SVG-generation prompt;
\item \textbf{\method{}}, evaluated across layered SVG, diagrams.net-compatible XML, and editable PPTX.
\end{itemize}

The Direct MLLM baseline receives the same raster image and canvas specification but generates SVG in a single direct process without category routing, representation selection, relation-aware SAR assembly, stage-local verification, or deterministic intermediate-state compilation.

\subsection{Supplementary Baselines}

We additionally include \textbf{Potrace} as a contour-tracing reference and \textbf{StarVector} as a general multimodal SVG-generation reference. These methods provide useful appearance-oriented comparisons but are not assumed to support all native authoring operations. Unsupported editing operations are scored as failures rather than omitted.

Each method is evaluated in the native representation officially supported or recommended by its authors. Format-specific audit parsers expose structures that already exist in the artifact; they do not repair, convert, or semantically reinterpret outputs.

For methods supporting multiple native formats, each format is evaluated independently and scores are averaged across all supported formats according to Eq.~\ref{eq:format_average}. We do not select the best format for a sample or operation. Human participants use the recommended authoring environment for the provided native artifact.

\subsection{All-Category and Common-Domain Protocols}

The all-category evaluation measures both reconstruction quality and category coverage. When a method does not officially support a benchmark category, the corresponding samples are counted as failures. The common-domain evaluation separately measures quality on scientific diagrams and academic posters, which are supported by all primary task-aligned methods.

For an unsupported or invalid output, VLM-Fid., Auto Edit-SR, Rel-SR, Human Edit-SR, and Ease receive zero; Text-CER and RFC receive one; LPIPS is computed against a blank canvas of the requested size; and output validity is zero. Such samples are never silently removed.

\subsection{Input Normalization}

All methods receive:

\begin{itemize}
\item the same raster image;
\item the same target canvas width and height;
\item the same aspect-ratio information;
\item no original source file;
\item no benchmark-specific object or relation annotations;
\item no manual correction.
\end{itemize}

Images are not selectively cropped, enhanced, or denoised for individual methods. When a method requires a fixed input resolution, the image is resized using its officially recommended preprocessing while preserving aspect ratio.

\subsection{Direct MLLM Prompt}

The Direct MLLM baseline uses the following fixed prompt:

\begin{quote}
\small
Given the input raster graphic, reconstruct it as an editable SVG with the same canvas size and visual appearance. Preserve all visible content, layout, text, formulas, colors, shapes, connectors, charts, tables, icons, and image regions. Represent recoverable elements as independently editable native SVG objects whenever possible. Do not replace the entire input with a single embedded image. Return only valid SVG code.
\end{quote}

The model version, decoding parameters, maximum tokens, retry policy, and image preprocessing are fixed for all samples.

\begin{table*}[t]
\centering
\small
\setlength{\tabcolsep}{3.5pt}
\renewcommand{\arraystretch}{1.18}

\begin{tabular}{
p{0.13\textwidth}|
p{0.20\textwidth}|
p{0.15\textwidth}|
p{0.34\textwidth}|
c
}
\hline
\multicolumn{1}{c|}{
\rule{0pt}{2.8ex}{\normalsize\bfseries Method}
}
&
\multicolumn{1}{c|}{
{\normalsize\bfseries Version}
}
&
\multicolumn{1}{c|}{
{\normalsize\bfseries Native Outputs}
}
&
\multicolumn{1}{c|}{
{\normalsize\bfseries Inference Configuration}
}
&
\multicolumn{1}{c}{
{\normalsize\bfseries \# Runs}
}
\\
\hline
Edit-Banana
&
official repository, July 2026 checkout;
\newline
GPT-5.5 (July 2026 API snapshot)
&
diagrams.net XML
&
official reconstruction pipeline;
GPT-5.5 for multimodal reasoning;
recommended preprocessing;
maximum 3 retries
&
3
\\
AutoFigure-Edit
&
v1.1;
\newline
GPT-5.5 (July 2026 API snapshot)
&
SVG
&
official editable-SVG pipeline;
GPT-5.5 for all MLLM calls;
official prompt and decoding configuration;
maximum 3 retries
&
3
\\
CraftEditor
&
official repository, July 2026 checkout;
\newline
GPT-5.5 (July 2026 API snapshot)
&
SVG
&
official raster-to-SVG pipeline;
GPT-5.5 for all VLM calls;
default SAM3 grounding configuration;
maximum 3 retries
&
3
\\
Direct MLLM
&
GPT-5.5
\newline
(July 2026 API snapshot)
&
SVG
&
fixed direct-generation prompt;
temperature 0;
maximum 16,384 output tokens;
no iterative repair
&
3
\\
\method{}
&
July 2026 implementation;
\newline
GPT-5.5 and GPT-image-2
&
SVG;
\newline
diagrams.net XML;
\newline
PPTX
&
fixed prompts and schemas;
representation-aware recovery;
$T_{\max}=3$;
fixed verification thresholds;
maximum 3 retries per stage
&
3
\\
Potrace
&
v1.16
&
SVG
&
official default tracing configuration;
SVG backend;
no semantic post-processing
&
1
\\
\hline
\end{tabular}

\caption{
Baseline versions and inference configurations.
All MLLM-based methods use the same GPT-5.5 July 2026 API snapshot and are
evaluated over three independent runs. Potrace is deterministic and is
evaluated once. No per-sample best-run selection is performed.
}
\label{tab:baseline_configuration}
\end{table*}
\subsection{Repeated Generations}

For stochastic methods, we perform $K=[\text{2}]$ independent generations using publicly reported fixed seeds. We report the mean and standard deviation across generations. We do not select the best generation using test-set metrics. When one artifact is required for human evaluation, the run is determined by a seed fixed before evaluation.

\subsection{Invalid Outputs}
\label{app:invalid_outputs}

An artifact is invalid if it cannot be parsed, cannot be opened in its intended environment, cannot be rendered, contains no visible reconstruction, is truncated, or is structurally incomplete. Invalid artifacts receive zero automatic and human editability scores. For visual metrics, an invalid artifact is represented by a blank canvas at the requested size. RFC is assigned the conservative value one. Output-validity rate is reported separately.

\section{Standardized Rendering and Artifact Audit}
\label{app:rendering}

\subsection{Rendering Protocol}

All inputs and reconstructed artifacts are rendered using a fixed canvas size, color space, background policy, resolution, font collection, and antialiasing configuration. The renderer version, operating system, DPI, font fallback order, transparency handling, and image-resampling method are reported in the released configuration.

Missing fonts are handled through a fixed fallback list shared across methods. A renderer failure is treated as an invalid output rather than retried with method-specific settings. Evaluation renderings are produced before any edit and after every executable edit.

\subsection{Common Audit Interface}

Format-specific parsers expose a read-only audit interface containing:

\begin{itemize}
\item artifact validity and canvas geometry;
\item native text objects and text content;
\item vector primitives and paths;
\item embedded raster resources;
\item connectors and endpoint metadata;
\item chart and table structures when present;
\item groups, containers, parent--child structure, and relation metadata;
\item rendered bounding boxes and stable object identifiers.
\end{itemize}

The audit interface does not add missing structure or convert one representation into another. It only exposes structures already present in the generated artifact.

\section{Evaluation Metrics}
\label{app:metrics}

\subsection{Notation and Aggregation}

Let
\begin{equation}
\mathcal{D}=\{I_i\}_{i=1}^{N}
\end{equation}
denote the evaluation images, and let $A_{i,m,f}$ denote the artifact generated by method $m$ for image $I_i$ in format $f$. Its standardized rendering is
\begin{equation}
R_{i,m,f}=\mathcal{R}(A_{i,m,f}).
\end{equation}

Unless stated otherwise, scores are first averaged over supported native formats without per-sample selection, then macro-averaged across the five benchmark categories. For a per-sample score $q_{i,m}$,
\begin{equation}
\overline{q}_m=
\frac{1}{C}
\sum_{c=1}^{C}
\frac{1}{|\mathcal{D}_c|}
\sum_{i\in\mathcal{D}_c}q_{i,m},
\qquad C=5.
\label{eq:macro_average}
\end{equation}
Micro-averaged and per-category results are reported in supplementary tables.

\subsection{Output Validity}

Let $v_{i,m}=1$ when the generated artifact is parseable, openable, and renderable, and zero otherwise. Output validity is
\begin{equation}
\mathrm{Valid}(m)=
\frac{1}{N}\sum_{i=1}^{N}v_{i,m}.
\label{eq:validity}
\end{equation}

\subsection{Perceptual Fidelity}

LPIPS is computed between the input raster and standardized rendering:
\begin{equation}
\mathrm{LPIPS}(m)=
\frac{1}{N}
\sum_{i=1}^{N}
d_{\mathrm{LPIPS}}(I_i,R_{i,m}).
\label{eq:lpips}
\end{equation}
Both images are resized using the same evaluation procedure. Lower LPIPS indicates greater perceptual similarity.

\subsection{VLM-Based Fidelity}

A fixed vision-language judge receives the input raster and reconstruction in randomized left--right order without method identity. It scores four dimensions from 1 to 5:

\begin{enumerate}
\item visible-content completeness;
\item layout and geometric consistency;
\item color, style, and appearance consistency;
\item text and formula legibility.
\end{enumerate}

The judge prompt is:

\begin{quote}
\small
Compare the reference raster graphic and the reconstructed rendering. Evaluate only visible reconstruction fidelity; do not infer or reward hidden editability. Score content completeness, layout and geometry, appearance and style, and text/formula legibility from 1 to 5. Return the four scores in the required structured format.
\end{quote}

Let $s_{i,m,j,d}\in\{1,\ldots,5\}$ denote judge call $j$ on dimension $d$. The score is normalized to $[0,1]$:
\begin{equation}
\begin{aligned}
\mathrm{VLM\text{-}Fid}(m)
&=
\frac{1}{4}
\left(
\bar{s}_m-1
\right),\\
\bar{s}_m
&=
\frac{1}{NJD}
\sum_{i=1}^{N}
\sum_{j=1}^{J}
\sum_{d=1}^{D}s_{i,m,j,d}.
\end{aligned}
\label{eq:vlm_fid}
\end{equation}
Here $D=4$. We use $J=[\text{3}]$ independent judge calls with randomized image order. Judge model, snapshot, prompt, temperature, and aggregation are fixed before final evaluation.

\subsection{Textual Consistency}

The same OCR system is applied to the input and reconstructed rendering:
\begin{equation}
t_i=\mathrm{OCR}(I_i),
\qquad
\hat{t}_{i,m}=\mathrm{OCR}(R_{i,m}).
\end{equation}
Low-confidence tokens are removed using a fixed threshold, and whitespace, Unicode variants, and punctuation are normalized. Text-CER is
\begin{equation}
\mathrm{Text\text{-}CER}(m)=
\frac{1}{N}
\sum_{i=1}^{N}
\frac{
\operatorname{Lev}(t_i,\hat{t}_{i,m})
}{
\max(1,|t_i|)
},
\label{eq:text_cer}
\end{equation}
where $\operatorname{Lev}$ is Levenshtein edit distance. Text-CER measures consistency with visible text recognized in the input, not absolute transcription accuracy.

We additionally report token-level precision, recall, and F1 as supplementary diagnostics.

\subsection{Raster Fallback Coverage}
\label{app:rfc}

RFC measures the fraction of visible input foreground covered by embedded raster elements in the output. Raster resources include SVG \texttt{image} elements and bitmap backgrounds, diagrams.net image cells, and PPTX picture shapes or raster fills.

Raster elements are rendered alone on a transparent canvas to obtain
\begin{equation}
M^{\mathrm{ras}}_{i,m}(x)\in\{0,1\}.
\end{equation}
The dominant input background color $b_i$ is estimated from border pixels, and the source foreground mask is
\begin{equation}
M^{\mathrm{fg}}_i(x)=
\mathbf{1}
\left[
\lVert I_i(x)-b_i\rVert_2>\tau_{\mathrm{fg}}
\right].
\end{equation}
Sample-level RFC is
\begin{equation}
\mathrm{RFC}_{i,m}=
\frac{
\sum_x M^{\mathrm{ras}}_{i,m}(x)M^{\mathrm{fg}}_i(x)
}{
\max\left(1,\sum_x M^{\mathrm{fg}}_i(x)\right)
}.
\label{eq:rfc}
\end{equation}
Overlapping raster patches are counted through their union. A full-canvas screenshot therefore yields RFC close to one even if editable elements are overlaid.

RFC is reported primarily on scientific diagrams, charts, and tables. For academic and design posters, raw raster occupancy is reported separately because independent raster panels may be semantically appropriate.

\subsection{Primitive Fragmentation Diagnostics}

As supplementary diagnostics, we report low-level primitive density and path complexity:
\begin{equation}
\mathrm{PrimitiveDensity}_{i,m}=
\frac{
N^{\mathrm{primitive}}_{i,m}
}{
A^{\mathrm{fg}}_i/10^6
},
\end{equation}
\begin{equation}
\mathrm{PathComplexity}_{i,m}=
\frac{
N^{\mathrm{path\ nodes}}_{i,m}
}{
\max(1,N^{\mathrm{path}}_{i,m})
}.
\end{equation}
These quantities may reveal contour-traced text or fragmented shapes, but are not treated as direct semantic-quality metrics because legitimate illustrations may require many primitives.

\section{Executable Native Editability Evaluation}
\label{app:executable_editability}

\subsection{Motivation}

Selectable layers do not necessarily imply meaningful editability. Text converted into paths cannot be edited as text, a rasterized arrow cannot expose marker attributes, and a flattened chart cannot support data updates. We therefore evaluate editability through executable authoring operations.

\subsection{Operation Applicability}

For operation $o$, a source-image detector produces candidates
\begin{equation}
\mathcal{C}_{i,o}=
\{(c_k,b_k,\gamma_k)\}_{k=1}^{K},
\end{equation}
where $c_k$ is candidate type, $b_k$ source region, and $\gamma_k$ confidence. The operation is applicable when
\begin{equation}
a_{i,o}=
\mathbf{1}\left[\max_k\gamma_k\geq\tau_o\right].
\label{eq:operation_applicability}
\end{equation}
If no sufficiently confident target exists, the operation is excluded from that sample's denominator.

Source detectors include OCR for text, formula-region detection, uniform-region and contour detection for shapes, line and marker detection for connectors, table and cell detection, chart and mark detection, and image-panel detection.

\subsection{Applicability-Detector Validation}

Detector thresholds are selected on a disjoint development set and fixed before final evaluation. On a manually audited subset, we report precision, recall, and agreement with human applicability judgments for every operation type, together with the fraction of candidates excluded by confidence filtering. Detectors operate only on source images and never inspect method outputs.

\subsection{Matching Source Targets to Native Objects}

For an applicable source target $b_{i,o}$, the format-specific parser exposes compatible native objects
\begin{equation}
\mathcal{P}_{i,m,o}=\{p_1,\ldots,p_K\}.
\end{equation}
Each object has rendered bounding box $B(p)$, native type $T(p)$, and optional content $C(p)$. The matching score is
\begin{align}
S(p,b_{i,o})={}&
\lambda_{\mathrm{box}}
\operatorname{IoU}(B(p),b_{i,o})
+
\lambda_{\mathrm{type}}
\mathbf{1}[T(p)\in\mathcal{T}_o]
\nonumber\\
&+
\lambda_{\mathrm{text}}
\operatorname{Sim}_{\mathrm{text}}(C(p),C_{i,o}),
\label{eq:target_matching}
\end{align}
where the text term is used only for content-dependent operations. The matched target is
\begin{equation}
p^\star=
\arg\max_{p\in\mathcal{P}_{i,m,o}}S(p,b_{i,o}).
\end{equation}
A match is valid only if the object satisfies operation-specific compatibility and exceeds the fixed matching threshold.

\subsection{Operation Suite}

\begin{table*}[t]
\centering
\scriptsize
\setlength{\tabcolsep}{3.0pt}
\renewcommand{\arraystretch}{1.13}
\resizebox{\textwidth}{!}{%
\begin{tabular}{l|l|l|l}
\toprule
\textbf{Operation}
& \textbf{Applicable Target}
& \textbf{Modification}
& \textbf{Required Native Capability}
\\
\midrule
Text replacement
& detected text
& replace with length-matched text
& editable text content
\\
Formula editing
& detected formula
& replace one symbol or number
& symbolic or semantic formula payload
\\
Fill modification
& filled shape or region
& change fill to fixed contrast color
& editable fill attribute
\\
Stroke modification
& vector shape or line
& change width or style
& editable stroke attribute
\\
Object resizing
& shape, icon, or panel
& scale by a fixed factor
& editable geometry
\\
Object translation
& shape, icon, or panel
& translate by fixed offset
& independent positioning
\\
Arrowhead editing
& connector or arrow
& add, remove, or change marker
& editable line/connector marker
\\
Table-cell editing
& detected table
& replace one cell value
& cell-level table structure
\\
Chart-data editing
& detected chart
& modify one numeric value
& data-backed or re-encodable chart
\\
Image-panel editing
& raster panel
& move or resize panel
& independent raster object
\\
\bottomrule
\end{tabular}
}
\caption{Executable object-level editing operations.}
\label{tab:edit_operations}
\end{table*}

\paragraph{Text replacement.}
The target string is replaced with a deterministic string of approximately equal length to reduce incidental layout changes.

\paragraph{Formula editing.}
A visible variable or numeric token is replaced by a deterministic alternative. Success requires a semantic formula payload, valid serialization, and a localized visual update.

\paragraph{Fill and stroke editing.}
The target attribute is changed to a fixed contrastive value derived from the original luminance and saturation.

\paragraph{Geometry editing.}
The target is resized by $1+\delta_s$ around its center or translated by a fixed fraction $\delta_t$ of the canvas dimension.

\paragraph{Arrowhead editing.}
The marker style is toggled or replaced with a predefined marker. Attachment preservation is evaluated separately by Rel-SR.

\paragraph{Table-cell editing.}
A detected non-header cell is selected and its text or numerical value is replaced. Replacing an entire table image does not satisfy the operation.

\paragraph{Chart-data editing.}
One selected value is multiplied by $1+\delta_c$, and the corresponding visual mark is regenerated. A chart represented only as an image or unrelated paths fails.

\paragraph{Image-panel editing.}
An intrinsic raster panel succeeds when it can be moved or resized independently without altering unrelated content.

\subsection{Format-Specific Operation Adapters}

SVG adapters modify existing XML element content, attributes, transforms, metadata, and group structure. diagrams.net adapters modify cell values, geometry, styles, parent structure, and edge metadata. PPTX adapters modify text frames, shape properties, transforms, tables, chart data, connectors, and pictures through Office Open XML structures.

Adapters may manipulate only structures already present in the artifact. They may not invoke another generative model, reconstruct a missing element, or replace raster content with new vector structure.

\subsection{Success Criteria}

Let $A_{i,m}$ and $A'_{i,m,o}$ be the artifacts before and after operation $o$, with renderings $R_{i,m}$ and $R'_{i,m,o}$. An edit succeeds only if:

\begin{enumerate}
\item a compatible native target is available;
\item the modified artifact can be saved and reparsed;
\item the modified artifact remains openable and renderable;
\item the requested content, style, geometry, or data attribute changes to the specified value;
\item the visual effect is localized to the intended region.
\end{enumerate}

Let $\Omega_{i,o}$ be the dilated union of the target region before and after editing:
\begin{equation}
D_{\mathrm{target}}=
\frac{1}{|\Omega_{i,o}|}
\sum_{x\in\Omega_{i,o}}
\lVert R'_{i,m,o}(x)-R_{i,m}(x)\rVert_1,
\end{equation}
\begin{equation}
D_{\mathrm{outside}}=
\frac{1}{|\overline{\Omega}_{i,o}|}
\sum_{x\notin\Omega_{i,o}}
\lVert R'_{i,m,o}(x)-R_{i,m}(x)\rVert_1.
\end{equation}
The visual check succeeds when
\begin{equation}
D_{\mathrm{target}}\geq\tau_{\mathrm{change}},
\qquad
D_{\mathrm{outside}}\leq\tau_{\mathrm{leak}}.
\end{equation}
Operation-specific structural assertions are also required. Let $s_{i,m,o}\in\{0,1\}$ denote final success.

\subsection{Automatic Edit Success Rate}

For operation $o$,
\begin{equation}
\mathrm{Edit\text{-}SR}_{m,o}=
\frac{
\sum_i a_{i,o}s_{i,m,o}
}{
\max(1,\sum_i a_{i,o})
}.
\label{eq:edit_sr_operation}
\end{equation}
The headline score is macro-averaged over operation types:
\begin{equation}
\mathrm{Edit\text{-}SR}_m=
\frac{1}{|\mathcal{O}|}
\sum_{o\in\mathcal{O}}
\mathrm{Edit\text{-}SR}_{m,o}.
\label{eq:edit_sr}
\end{equation}

Let $n^{\mathrm{app}}_{m,o}$, $n^{\mathrm{match}}_{m,o}$,
$n^{\mathrm{write}}_{m,o}$, and $n^{\mathrm{local}}_{m,o}$ denote the
numbers of applicable targets, native matches, valid writes, and localized
successful changes. We additionally report
\begin{equation}
\mathrm{NativeMatchRate}_{m,o}=
\frac{n^{\mathrm{match}}_{m,o}}
{\max(1,n^{\mathrm{app}}_{m,o})},
\end{equation}
\begin{equation}
\mathrm{ValidWriteRate}_{m,o}=
\frac{n^{\mathrm{write}}_{m,o}}
{\max(1,n^{\mathrm{match}}_{m,o})},
\end{equation}
\begin{equation}
\mathrm{LocalizedEffectRate}_{m,o}=
\frac{n^{\mathrm{local}}_{m,o}}
{\max(1,n^{\mathrm{write}}_{m,o})}.
\end{equation}

\section{Relation-Preserving Editability Evaluation}
\label{app:relation_editability}

\subsection{Scope and Relation Targets}

Rel-SR evaluates attachment, grouping, and containment because these relations have observable consequences under editing. A flattened image does not uniquely determine an exhaustive authoring graph, so the relation-evaluation set uses sparse operation-level relation records rather than complete object-tree annotation.

Candidate relations are proposed from the source image using structural and geometric detectors, then verified by annotators without inspecting method outputs. Each record contains the relation type, participating source regions, prescribed perturbation, and evaluation tolerance. Let $\mathcal{G}_{\rho}$ denote verified instances of relation type
\begin{equation}
\rho\in
\{\mathrm{attach},\mathrm{group},\mathrm{contain}\}.
\end{equation}

\subsection{Relation-Preserving Operations}

\paragraph{Attachment.}
A source or target object is translated or resized. The connector must remain attached to the same object, and the transformed endpoint must remain within the predefined boundary or anchor tolerance.

\paragraph{Grouping.}
A group is translated or resized. All member objects must remain members of the same group and preserve the prescribed relative transform under the group operation.

\paragraph{Containment.}
A container is translated or resized. The contained object must remain associated with the same container and remain inside the transformed containment region up to the fixed tolerance.

The operation adapter may use only native relations, group structure, parent structure, or relation metadata already present in the artifact. It may not add a missing relation after generation.

\subsection{Native Relation Matching}

Source relation participants are matched to compatible output objects using the same spatial, type, and content evidence as object-level matching. A relation match exists only when all required participants are matched and the output contains compatible relation structure or relation metadata. Visual proximity alone before editing is insufficient.

\subsection{Success Criteria}

For relation instance $g\in\mathcal{G}_{\rho}$, let $r_{g,m,\rho}\in\{0,1\}$. Success requires:

\begin{enumerate}
\item compatible native participating objects are found;
\item the prescribed transformation is serialized and rendered;
\item the requested target transformation occurs;
\item the same operational relation remains structurally represented when the format exposes relation metadata;
\item the relation remains geometrically valid after editing.
\end{enumerate}

For SVG backends without native editor-level endpoint references, compiler-emitted relation metadata and the corresponding native SVG operation adapter are used to apply and verify the prescribed relation-dependent transformation. Such metadata must already exist before evaluation.

\subsection{Relation-Preserving Edit Success Rate}

For relation type $\rho$,
\begin{equation}
\mathrm{Rel\text{-}SR}_{m,\rho}=
\frac{
\sum_{g\in\mathcal{G}_{\rho}}r_{g,m,\rho}
}{
\max(1,|\mathcal{G}_{\rho}|)
}.
\label{eq:relation_type_sr}
\end{equation}
The headline score is macro-averaged:
\begin{equation}
\mathrm{Rel\text{-}SR}_m=
\frac{1}{3}
\sum_{\rho\in
\{\mathrm{attach},\mathrm{group},\mathrm{contain}\}}
\mathrm{Rel\text{-}SR}_{m,\rho}.
\label{eq:relation_sr}
\end{equation}

\begin{table*}[t]
\centering
\caption{
Relation-preserving edit success by relation type.
Attachment measures whether connector endpoints remain attached after object
transformations; Grouping measures whether group membership and relative
transforms are preserved; Containment measures whether contained objects remain
spatially valid within their containers. Rel-SR denotes the macro-average over
the three relation types.
}
\label{tab:relation_results}

\vspace{2pt}

\fontsize{9.5pt}{11.5pt}\selectfont
\renewcommand{\arraystretch}{1.15}

\begin{tabular*}{0.78\textwidth}{
    @{\extracolsep{\fill}}
    l
    c
    c
    c
    c
    @{}
}
\toprule

\textbf{Method}
& \textbf{Attachment} $\uparrow$
& \textbf{Grouping} $\uparrow$
& \textbf{Containment} $\uparrow$
& \textbf{Rel-SR} $\uparrow$
\\

\midrule

Edit-Banana
& 0.61
& 0.53
& 0.45
& 0.53
\\

AutoFigure-Edit
& 0.48
& 0.42
& 0.36
& 0.42
\\

CraftEditor
& 0.72
& 0.65
& 0.55
& 0.64
\\

Direct MLLM
& 0.29
& 0.24
& 0.19
& 0.24
\\

\addlinespace[2pt]

\textbf{\method{}}
& \textbf{0.94}
& \textbf{0.89}
& \textbf{0.81}
& \textbf{0.88}
\\

\bottomrule
\end{tabular*}

\end{table*}

\begin{table*}[t]
\centering
\caption{
Relation-wise ablation results.
Removing representation selection reduces the availability of
operation-compatible objects, while removing stage-local verification
leaves attachment, grouping, and containment inconsistencies unresolved.
}
\fontsize{9.5pt}{11.5pt}\selectfont
\renewcommand{\arraystretch}{1.15}

\begin{tabular*}{0.78\textwidth}{
    @{\extracolsep{\fill}}
    l
    c
    c
    c
    c
    @{}
}
\toprule

\textbf{Configuration}
& \textbf{Attachment} $\uparrow$
& \textbf{Grouping} $\uparrow$
& \textbf{Containment} $\uparrow$
& \textbf{Rel-SR} $\uparrow$
\\

\midrule

\textbf{Full \method{}}
& \textbf{0.94}
& \textbf{0.89}
& \textbf{0.81}
& \textbf{0.88}
\\

\midrule

w/o Representation Selection
& 0.82
& 0.73
& 0.64
& 0.73
\\

w/o Stage-Local Verification
& 0.72
& 0.64
& 0.56
& 0.64
\\

\bottomrule
\end{tabular*}

\label{tab:relation_ablation}

\end{table*}

\section{Human Evaluation}
\label{app:human_study}

\subsection{Objectives}

The human study evaluates perceived visual fidelity, successful completion of practical editing tasks, and perceived editing ease. Completion time is recorded as a secondary end-to-end usability diagnostic because methods may use different authoring environments.

\subsection{Participants}

We recruit $P=[\text{10}]$ participants with prior experience in at least one visual-authoring tool, such as PowerPoint, diagrams.net, Illustrator, Inkscape, or an equivalent editor. We report editing experience, familiarity with each environment, compensation, and the institutional-review or exemption procedure when applicable.

Before the formal study, each participant completes a tutorial and practice tasks excluded from evaluation.

\subsection{Sample and Task Selection}

The study subset is stratified across the five benchmark categories. Each sample is paired with one object-level or relation-dependent editing instruction generated from the fixed operation suite. The interface displays the original raster, reconstructed artifact, highlighted target, and concise instruction.

Example tasks include replacing a label, editing a formula token, changing shape fill, resizing a shape, changing a connector marker, updating a table cell, changing chart data, moving an image panel, or moving an object while preserving an attached connector.

\subsection{Experimental Design}

We use a blinded and balanced assignment:

\begin{itemize}
\item method names are hidden;
\item artifact and task order are randomized;
\item no participant evaluates multiple methods on the same source image;
\item methods and categories are balanced across participants;
\item the assignment follows a balanced incomplete-block design.
\end{itemize}

Participants use each method's recommended native authoring environment. For every environment, we report software version, operating system, plugin configuration, permitted actions, tutorial duration, keyboard-shortcut policy, and task timeout.

\subsection{Human Fidelity}

Before editing, participants rate how faithfully the reconstruction preserves visible content, layout, appearance, and element completeness on a five-point scale. The headline score is normalized to $[0,1]$:
\begin{equation}
\mathrm{HumanFid}(m)=
\frac{1}{|\mathcal{H}_m|}
\sum_{h\in\mathcal{H}_m}
\frac{r^{\mathrm{fid}}_h-1}{4}.
\label{eq:human_fid}
\end{equation}
Raw 1--5 means and rating distributions are additionally reported.

\subsection{Human Editing Success}

A task succeeds when the requested edit is completed and the submitted file satisfies its predefined structural and rendering conditions:
\begin{equation}
\mathrm{HumanEdit\text{-}SR}(m)=
\frac{
\#\text{successfully completed tasks}
}{
\max(1,\#\text{assigned tasks})
}.
\label{eq:human_edit_sr}
\end{equation}
Automatic checkers are used whenever possible. Remaining cases are reviewed by evaluators blinded to method identity.

\subsection{Editing Ease}

After each task, participants rate ease from 1 to 5, where one means impossible or requiring reconstruction and five means direct and effortless. The headline score is
\begin{equation}
\mathrm{Ease}(m)=
\frac{1}{|\mathcal{H}_m|}
\sum_{h\in\mathcal{H}_m}
\frac{r^{\mathrm{ease}}_h-1}{4}.
\label{eq:human_ease}
\end{equation}
We additionally report raw means and rating distributions.

\subsection{Completion Time}

Timing begins after the artifact loads and the instruction appears, and ends when the edited artifact is submitted. A timeout is a failure. For successful tasks,
\begin{equation}
\mathrm{MedTime}(m)=
\operatorname{median}\{t_h\mid s_h=1\}.
\end{equation}
We also report capped time:
\begin{equation}
t_h^{\mathrm{cap}}=\min(t_h,t_{\max}),
\end{equation}
\begin{equation}
\mathrm{CappedTime}(m)=
\frac{1}{|\mathcal{H}_m|}
\sum_{h\in\mathcal{H}_m}t_h^{\mathrm{cap}}.
\end{equation}

\subsection{Quality Control and Agreement}
Trials are excluded only when the participant does not open the provided artifact, does not attempt the assigned task, encounters a documented system failure unrelated to the artifact, or violates the study instructions. All exclusion criteria are fixed before data collection.

Inter-rater agreement is assessed separately for visual fidelity and editing ease using Krippendorff's $\alpha$ with an ordinal distance function, which accommodates the incomplete assignment design and the ordinal nature of the five-point ratings. We report $\alpha_{\mathrm{fid}}=[\text{TO FILL}]$ for fidelity and $\alpha_{\mathrm{ease}}=[\text{TO FILL}]$ for editing ease. The corresponding 95

\section{Controlled Output-Format Analysis}
\label{app:controlled_format}

The primary evaluation uses each method's officially supported native representation. To reduce editor-specific confounding, we additionally conduct a controlled human study for methods that natively support SVG. All included methods use the same SVG editor, version, tutorial, task set, rendering configuration, and success criteria.

Methods without native SVG support are not converted to SVG because conversion could independently add or remove editability.

\begin{table}[t]
\centering
\caption{Controlled human editing results in a common SVG environment.}
\label{tab:svg_controlled}

\vspace{1pt}

\fontsize{8.8pt}{10.5pt}\selectfont
\renewcommand{\arraystretch}{1.12}

\begin{tabular*}{0.94\columnwidth}{
    @{\extracolsep{\fill}}
    l
    c
    c
    c
    @{}
}
\toprule

\textbf{Method}
& \textbf{Edit-SR} $\uparrow$
& \textbf{Ease} $\uparrow$
& \textbf{Time} $\downarrow$
\\

\midrule

AutoFigure-Edit
& 0.36
& 0.41
& 0.34
\\

CraftEditor
& 0.45
& 0.52
& 0.67
\\

Direct MLLM
& 0.17
& 0.06
& 0.93
\\

\midrule

\textbf{\method{}-SVG}
& \textbf{0.57}
& \textbf{0.64}
& \textbf{0.26}
\\

\bottomrule
\end{tabular*}

\end{table}

\section{Multi-Format Export Analysis}
\label{app:multiformat}

\method{} compiles the same SAR into layered SVG, diagrams.net-compatible XML, and editable PPTX. We evaluate validity, rendering fidelity, editability, relation preservation, and file size separately for each backend.

For format $f$, let $N_f^{\mathrm{valid}}$ denote the number of
successfully opened and rendered exports. Export validity is
\begin{equation}
\mathrm{ExportValid}(f)=
\frac{N_f^{\mathrm{valid}}}{N}.
\end{equation}

Let $R_i^f$ be the standardized rendering from format $f$. Cross-format consistency is
\begin{equation}
\begin{aligned}
\mathrm{CrossFormatLPIPS}
&=
\frac{1}{N\binom{|\mathcal{F}|}{2}}
\sum_{i=1}^{N}D_i,\\
D_i
&=
\sum_{\substack{f_a,f_b\in\mathcal{F}\\f_a<f_b}}
d_{\mathrm{LPIPS}}(R_i^{f_a},R_i^{f_b}).
\end{aligned}
\label{eq:cross_format}
\end{equation}

\begin{table*}[t]
\centering
\fontsize{9.3pt}{11.2pt}\selectfont
\renewcommand{\arraystretch}{1.15}

\begin{tabular*}{0.90\textwidth}{
    @{\extracolsep{\fill}}
    lccccc
    @{}
}
\toprule
\textbf{Format}
& \textbf{Export Valid} $\uparrow$
& \textbf{VLM-Fid.} $\uparrow$
& \textbf{Edit-SR} $\uparrow$
& \textbf{Rel-SR} $\uparrow$
& \textbf{Median Size (MB)} $\downarrow$
\\
\midrule

Layered SVG
& \textbf{0.99}
& \textbf{0.78}
& 0.78
& 0.84
& 1.67
\\

diagrams.net XML
& 0.98
& 0.74
& \textbf{0.81}
& \textbf{0.93}
& \textbf{1.43}
\\

Editable PPTX
& 0.97
& 0.76
& 0.78
& 0.87
& 2.67
\\

\bottomrule
\end{tabular*}

\caption{
Validity, fidelity, editability, and relation preservation across
\method{} output formats.
Export Valid denotes the fraction of artifacts that can be successfully
opened and rendered.
VLM-Fid., Edit-SR, and Rel-SR are evaluated independently for each format
using the same protocol as the main evaluation.
Median Size denotes the median file size of valid artifacts.
The mean scores across SVG, diagrams.net XML, and PPTX correspond to the
multi-format \method{} results reported in Table~\ref{tab:main_results}.
}
\label{tab:multiformat}
\end{table*}

\subsection{Format-Specific Limitations}

SVG does not provide universal editor-level semantics for every attachment or chart relation, so the compiler stores structured metadata and groups when a direct primitive is unavailable. diagrams.net provides explicit graph edges and parent--child cells but offers more limited typography and chart primitives. PPTX provides native text, shapes, tables, charts, pictures, and groups, but some relation and formula semantics depend on Office Open XML support. Any approximation is recorded in the export log and evaluated through the same format-specific checkers.

\section{Ablation Studies}
\label{app:ablations}

\subsection{Representation Selection}

The \textbf{w/o Representation Selection} variant replaces type-specific representation decisions with a generic recovery strategy. Routing, evidence parsing, relation assembly, verification budget, and compilation remain unchanged. This variant may reproduce local appearance using raster patches, low-level paths, or flattened structures but loses operation-compatible payloads.

\subsection{Stage-Local Verification}

The \textbf{w/o Stage-Local Verification} variant performs parsing, recovery, and assembly without targeted verification or repair. Routing, representation selection, SAR, and compilers remain unchanged. This ablation tests whether localized correction resolves object, relation, rendering, and layout errors without regenerating validated content.

\subsection{Relation-Aware Assembly}

The supplementary \textbf{w/o Relation-Aware Assembly} variant retains recovered visual objects but removes explicit relation resolution. Connectors, groups, and containers are serialized as independent visual elements whenever possible. This variant directly tests the source of Rel-SR improvements.

\subsection{Number of Verification Rounds}

We evaluate
\begin{equation}
T\in\{0,1,\ldots,T_{\max}\}.
\end{equation}
For each budget, we report fidelity, Edit-SR, Rel-SR, latency, token usage, and cost.

\section{Efficiency and Resource Usage}
\label{app:efficiency}

For sample $i$, end-to-end latency is decomposed as
\begin{equation}
T_i=
T_i^{\mathrm{route}}
+
T_i^{\mathrm{parse}}
+
T_i^{\mathrm{recover}}
+
T_i^{\mathrm{verify}}
+
T_i^{\mathrm{export}}.
\end{equation}
Average latency is
\begin{equation}
\overline{T}=
\frac{1}{N}\sum_{i=1}^{N}T_i.
\end{equation}
Total token consumption is
\begin{equation}
\mathrm{Tokens}=
\sum_{i=1}^{N}
\left(
n_i^{\mathrm{input}}+n_i^{\mathrm{output}}
\right),
\end{equation}
and average API cost is
\begin{equation}
\mathrm{CostPerSample}=
\frac{\sum_{i=1}^{N}\mathrm{Cost}_i}{N}.
\end{equation}

Latency and memory are measured under a fixed hardware and software configuration. We report CPU, GPU, RAM, operating system, concurrency, API region when relevant, caching policy, retry handling, and whether network time and renderer startup are included.

\begin{table*}[t]
\centering
\fontsize{8.8pt}{10.5pt}\selectfont
\renewcommand{\arraystretch}{1.15}

\begin{tabular*}{0.98\textwidth}{
    @{\extracolsep{\fill}}
    l
    c
    r
    r
    r
    r
    c
    @{}
}
\toprule
\textbf{Input Category}
& \textbf{\# Samples}
& \shortstack{\textbf{Avg. Input}\\\textbf{Tokens}}
& \shortstack{\textbf{Avg. Output}\\\textbf{Tokens}}
& \shortstack{\textbf{Avg. Total}\\\textbf{Tokens}}
& \shortstack{\textbf{Aggregate}\\\textbf{Tokens}}
& \shortstack{\textbf{Avg. API}\\\textbf{Cost}}
\\
\midrule

Scientific diagrams
& 100
& 56,468
& 20,229
& 76,697
& 7,669,700
& \$2.50
\\

Academic posters
& 50
& 205,000
& 72,000
& 277,000
& 13,850,000
& \$9.00
\\

Design posters
& 50
& 160,000
& 55,000
& 215,000
& 10,750,000
& \$6.50
\\

Tables
& 25
& 14,800
& 5,200
& 20,000
& 500,000
& \$0.50
\\

Charts
& 25
& 12,500
& 4,800
& 17,300
& 432,500
& \$0.50
\\

\midrule

\textbf{Overall}
& \textbf{250}
& \textbf{98,317}
& \textbf{34,492}
& \textbf{132,809}
& \textbf{33,202,200}
& \textbf{\$4.20}
\\

\bottomrule
\end{tabular*}

\caption{
Token consumption and estimated API cost of \method{} across the five
input categories.
Per-sample token counts include all recorded GPT-5.5 calls during routing,
parsing, representation-aware recovery, relation assembly, and stage-local
verification.
API cost includes both GPT-5.5 and GPT-image-2 calls.
Because GPT-image-2 usage is not represented using the same text-token
accounting, API cost is not strictly proportional to the reported token counts.
The overall values are weighted by the number of samples in each category.
}
\label{tab:token_usage}
\end{table*}

Efficiency is additionally reported by category because posters generally contain more visible elements than individual charts, tables, or diagrams.

\section{Statistical Analysis}
\label{app:statistics}

For automatic metrics, we report category-macro means and 95\% confidence intervals obtained through bootstrap resampling over source images. For stochastic methods, we additionally report variation across fixed independent generations.

Pairwise method comparisons use paired bootstrap tests because methods are evaluated on the same source images. Multiple comparisons are corrected using the Holm--Bonferroni procedure. Statistical significance is evaluated at $\alpha=0.05$.

For human fidelity and ease, confidence intervals are bootstrapped over both participants and source images. Human editing success is binary and is analyzed using a mixed-effects logistic model or participant-level paired procedure selected before analysis, with participant and source image treated as random effects where applicable. We report absolute differences and confidence intervals in addition to $p$-values.

\section{Additional Qualitative Results}
\label{app:qualitative}

Supplementary qualitative results cover all five benchmark categories and include both typical successes and representative failures. Each example contains the input raster, baseline reconstructions, native object or layer structure, an artifact after a fixed editing operation, and a zoomed comparison of the edited target.

Examples are selected before inspecting per-method aggregate metric values. The analysis focuses on:

\begin{itemize}
\item missing visible elements;
\item unnecessary raster fallback;
\item traced or rasterized text;
\item corrupted or flattened formulas;
\item fragmented shapes and icons;
\item detached or rasterized connectors;
\item flattened charts and tables;
\item invalid grouping or containment;
\item unintended changes outside the edited region;
\item invalid or non-renderable outputs.
\end{itemize}

For \method{}, the supplementary examples also compare the same recovered SAR compiled into SVG, diagrams.net-compatible XML, and PPTX, including cases where a backend approximates unsupported semantics.

\begin{figure*}[htbp]
\centering
\includegraphics[width=\textwidth]{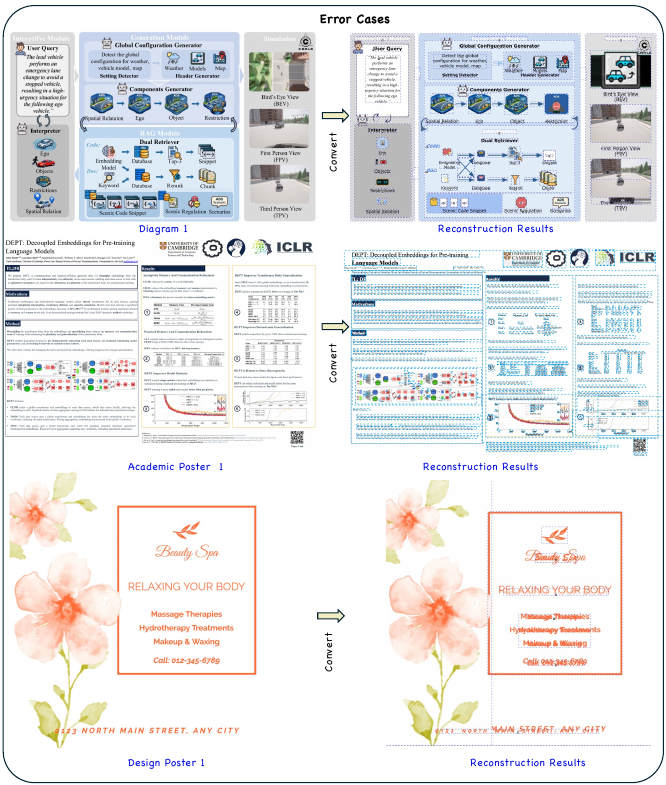}
\caption{
Qualitative reconstruction results for scientific diagrams. The outputs are opened in Inkscape, where blue selection outlines indicate independently editable objects and are not part of the rendered output. \method{} accurately reconstructs diagrams with complex nested layouts and conventional multi-box, multi-arrow structures while maintaining high visual fidelity.
}
\label{fig:error_case}
\end{figure*}

\section{Failure Cases and Limitations}
\label{app:failure_cases}
Representative failure cases are shown in Figure~\ref{fig:error_case}. These examples illustrate that the remaining errors mainly arise from insufficient visual evidence, ambiguous structural interpretation, and limitations of the target authoring formats, rather than from output serialization alone.

\paragraph{Small and degraded text.}
Extremely small text, JPEG artifacts, blur, or low contrast may cause OCR errors. Verification can correct some visible inconsistencies but cannot recover information absent from the raster input.

\paragraph{Mathematical expressions.}
Dense or heavily stylized formulas may be recognized incorrectly. A visually plausible expression may also differ from the intended symbolic semantics.

\paragraph{Dense connectors.}
Overlapping or closely spaced connectors may yield incorrect endpoint association, missing markers, or ambiguous routing.

\paragraph{Stylized shapes and icons.}
Highly decorative elements may not admit a compact parameterized representation. They may be reconstructed as grouped vector paths or localized raster content, reducing semantic editability.

\paragraph{Ambiguous representation selection.}
Some regions admit multiple reasonable representations. A decorative wordmark, for example, may be treated as typography, grouped vectors, or image-like content. The intended historical authoring choice is not observable from pixels alone.

\paragraph{Chart and table recovery.}
Low-resolution axes, dense labels, unconventional encodings, merged cells, or irregular layouts may prevent reliable recovery of data and topology.

\paragraph{Intrinsic raster content.}
Distinguishing appropriate raster panels from avoidable raster fallback is difficult in poster categories without exhaustive semantic annotation. We therefore interpret poster raster occupancy separately from vector-dominant RFC.

\paragraph{Format limitations.}
SVG, diagrams.net XML, and PPTX differ in available primitives and relation semantics. Some SAR content must be represented through structured compositions or metadata rather than a direct native counterpart.

\paragraph{Editor-dependent interaction.}
Human task time reflects both artifact structure and editor usability. We therefore treat task success and perceived ease as primary usability measures and time as secondary.

\paragraph{Computational cost.}
Multi-stage parsing, recovery, verification, and compilation require more calls than one-pass generation. Selective verification, parallel recovery, shared visual features, and confidence-based early stopping may improve efficiency.

\section{Ethical Considerations and Data Governance}
\label{app:ethics}

Benchmark metadata records source and usage status. Images are redistributed only when permitted; otherwise, the release provides identifiers or retrieval instructions. Samples containing sensitive personal information are excluded or appropriately redacted before evaluation.

For the human study, we report approval or exemption status, informed-consent procedure, compensation, withdrawal policy, and data-retention practice. Participant identities are not included in released results. Benchmark images are used only for reconstruction and editing evaluation.





\end{document}